\documentclass{article}
\usepackage{pdfsync}
\usepackage{amsfonts, amsmath, amsopn, amssymb, amsthm}
\usepackage{subfig}
\usepackage{bbm}
\usepackage{url}
\usepackage{algpseudocode,algorithm} 
\usepackage{graphicx}
\usepackage{epstopdf}
\usepackage{natbib}

\newcommand{\indicator}[1]{\mathbbm{1}{\left[ {#1} \right] }}

\newcommand{\No}{\mathcal{N}}
\newcommand{\bs}{\backslash}
\newcommand{\norm}[1]{\left\|#1\right\|}
\newcommand{\argmin}{\mathop{\rm argmin}}
\newcommand{\minimize}{\mathop{\rm minimize}}

\newtheorem{theorem}{Theorem}
 
\newtheorem{proposition}[theorem]{Proposition} 
\newtheorem{remark}[theorem]{Remark}

\begin{document}

\title{Learning the Structure of Mixed Graphical Models}


\author{Jason D. Lee\footnote{Institute of Computational and Mathematical Engineering, Stanford University.}
 and Trevor J. Hastie \footnote{Department of Statistics, Stanford University.}}
\maketitle

\begin{abstract}
  We consider the problem of learning the structure of a pairwise
  graphical model over continuous and discrete variables. We present a
  new pairwise model for graphical models with both continuous and
  discrete variables that is amenable to structure learning. In
  previous work, authors have considered structure learning of
  Gaussian graphical models and structure learning of discrete
  models. Our approach is a natural generalization of these two lines
  of work to the mixed case. The penalization scheme involves a novel
  symmetric use of the group-lasso norm and follows naturally from a
  particular parametrization of the model.
\end{abstract}



\section{Introduction}
Many authors have considered the problem of learning the edge
structure and parameters of sparse undirected graphical models. We
will focus on using the $l_1$ regularizer to promote sparsity.  This
line of work has taken two separate paths: one for learning continuous
valued data and one for learning discrete valued data.  However, typical
data sources contain both continuous and discrete
variables: population survey data, genomics data, url-click pairs etc. For genomics data, in addition to the gene expression values, we have attributes attached to each sample such as gender, age, ethniticy etc. In
this work, we consider learning mixed models with both continuous
variables and discrete variables.

For only continuous variables, previous work assumes a
multivariate Gaussian (Gaussian graphical) model with mean $0$ and
inverse covariance $\Theta$. $\Theta$ is then estimated via the
graphical lasso by minimizing the regularized negative log-likelihood
$\ell(\Theta)+\lambda \norm{\Theta}_{1}$. Several efficient methods
for solving this can be found in \cite{friedman2008,
  banerjee2008}. Because the graphical lasso problem is
computationally challenging, several authors considered methods
related to the pseudolikelihood (PL) and nodewise regression
\citep{meinshausen06, friedman2010,peng2009}.  For discrete models,
previous work focuses on estimating a pairwise Markov random field of
the form $p(y) \propto \exp{\sum_{r\leq j} \phi_{rj}(y_r,y_j)}$. The
maximum likelihood problem is intractable for models with a moderate
to large number of variables (high-dimensional) because it requires
evaluating the partition function and its derivatives. Again previous work has focused
on the pseudolikelihood approach 
\citep{guo2010joint,schmidt2010,schmidt2008,hoefling2009,jalali2011,lee2006,ravikumar2010}. 

Our main contribution here is to propose a model that connects the
discrete and continuous models previously discussed. The conditional
distributions of this model are two widely adopted and well understood
models: multiclass logistic regression and Gaussian linear
regression. In addition, in the case of only discrete variables, our
model is a pairwise Markov random field; in the case of only
continuous variables, it is a Gaussian graphical model. Our proposed
model leads to a natural scheme for structure learning that
generalizes the graphical Lasso. Here the parameters occur as
singletons, vectors or blocks, which we penalize using group-lasso
norms, in a way that respects the symmetry in the model. Since each
parameter block is of different size, we also derive a calibrated
weighting scheme to penalize each edge fairly. We also discuss a
conditional model (conditional random field) that allows the output
variables to be mixed, which can be viewed as a multivariate response
regression with mixed output variables. Similar ideas have been used
to learn the covariance structure in multivariate response regression
with continuous output variables
\cite{witten2009covariance,kim2009multivariate,rothman2010sparse}.

In Section \ref{sec:mgm}, we introduce our new mixed graphical model and discuss
previous approaches to modeling mixed data. Section \ref{sec:paramest} discusses the
pseudolikelihood approach to parameter estimation and connections to
generalized linear models. Section \ref{sec:penalty} discusses a natural method to
perform structure learning in the mixed model. Section \ref{sec:calibration} presents the
calibrated regularization scheme, Section \ref{sec:MSC} discusses the consistency of the estimation procedures, and Section \ref{sec:optalg} discusses two methods
for solving the optimization problem. Finally, Section \ref{sec:condmodel} discusses a
conditional random field extension and Section \ref{sec:exp} presents empirical
results on a census population survey dataset and synthetic
experiments.

\section{Mixed Graphical Model}
\label{sec:mgm}
We propose a pairwise graphical model on continuous and discrete variables. The model is a pairwise Markov random field with density $p(x,y;\Theta)$ proportional to
\begin{align}
\exp{\left(\sum_{s=1}^p\sum_{t=1}^p-\frac{1}{2} \beta_{st} x_{s} x_{t}+\sum_{s=1}^{p}\alpha_{s} x_{s} +\sum_{s=1}^p \sum_{j=1}^{q} \rho_{sj}(y_{j})x_{s}+\sum_{j=1}^q \sum_{r=1}^q \phi_{rj}(y_r , y_j)\right)}.
\label{eq:jointdensity}
\end{align}
Here $x_s$ denotes the $s$th of $p$ continuous variables, and $y_j$ the $j$th of $q$ discrete variables. 
The joint model is parametrized by $\Theta= [\{\beta_{st}\}, \{\alpha_s \}, \{\rho_{sj}\},\{\phi_{rj}\}]$\footnote{$\rho_{sj}(y_j )$ is a function taking $L_j$ values $\rho_{sj}(1),\ldots,\rho_{sj}(L_j)$. Similarly, $\phi_{rj}(y_r,y_j)$ is a bivariate function taking on $L_r \times L_j$ values. Later, we will think of $\rho_{sj}(y_j)$ as a vector of length $L_j$ and $\phi_{rj}(y_r,y_j)$ as a matrix of size $L_r \times L_j$.}. The discrete $y_r$ takes on $L_r$ states. The model parameters are $\beta_{st}$ continuous-continuous edge potential, $\alpha_{s}$ continuous node potential, $\rho_{sj}(y_{j})$ continuous-discrete edge potential, and $\phi_{rj}(y_r,y_j)$ discrete-discrete edge potential.

The two most important features of this model are:
\begin{enumerate}
\item 
the conditional distributions are given by Gaussian linear regression and multiclass logistic regressions;
\item the model simplifies to a multivariate Gaussian in the case of only continuous variables and simplifies to the usual discrete pairwise Markov random field in the case of only discrete variables.
\end{enumerate}
The conditional distributions of a graphical model are of critical
importance. The absence of an edge corresponds to two variables being
conditionally independent. The conditional independence can be read
off from the conditional distribution of a variable on all others. For
example in the multivariate Gaussian model, $x_s$ is conditionally
independent of $x_t$ iff the partial correlation coefficient is
$0$. The partial correlation coefficient is also the regression
coefficient of $x_t$ in the linear regression of $x_s$ on all other
variables. Thus the conditional independence structure is captured by
the conditional distributions via the regression coefficient of a
variable on all others. Our mixed model has the desirable property
that the two type of conditional distributions are simple Gaussian
linear regressions and multiclass logistic regressions. This follows
from the pairwise property in the joint distribution. In more detail:
\begin{enumerate}
\item The conditional distribution of $y_r$ given the rest is multinomial, with probabilities defined  by a multiclass logistic regression where the covariates are the other variables $x_s$ and $y_{\bs r}$ (denoted collectively by $z$ in the right-hand side):
  \begin{equation}
    \label{eq:simple1}
p(y_r =k|y_{\bs r}, x; \Theta) =\frac{\exp{\left( \omega_{k}^{T} z \right)}}{\sum_{l=1}^{L_r} \exp{\left( \omega_{l}^{T} z\right) }} = \frac{\exp{\left( \omega_{0k} + \sum_{j} \omega_{kj} z_j \right)}} {\sum_{l=1}^{L_r}\exp{\left( \omega_{0l} + \sum_j \omega_{lj} z_j \right)}}
\end{equation}
Here we use a simplified notation, which we make explicit in Section~\ref{sec:pseudolikelihood}. The discrete variables are represented as dummy variables for each state, e.g.  $z_j = \indicator{y_u = k}$, and for continuous variables $z_s =x_s$.
\item The conditional distribution of $x_s$ given the rest is Gaussian, with a mean function defined by a linear regression with predictors $x_{\bs s}
$ and $y_r$.
\begin{align}
E(x_s | x_{\bs s}, y_r;\Theta ) &=\omega^{T} z= \omega_0 +\sum_j z_j \omega_j\label{eq:simple2}\\
p(x_s | x_{\bs s}, y_r;\Theta )&= \frac{1}{\sqrt{2\pi} \sigma_s} \exp{\left(-\frac{1}{2 \sigma_s^2} ( x_{s} -\omega^{T} z )^2 \right)}.\nonumber
\end{align}
As before, the discrete variables are represented as dummy variables for each state $z_j = \indicator{y_u = k}$ and for continuous variables $z_s =x_s$.
\end{enumerate}
The exact form of the conditional distributions (\ref{eq:simple1}) and (\ref{eq:simple2}) are given in (\ref{eq:discond}) and (\ref{eq:ctscond}) in Section~\ref{sec:pseudolikelihood}, where the  regression parameters $\omega_j$ are defined in terms of the parameters $\Theta$.

The second important aspect of the mixed model is the two special cases of only continuous and only discrete variables.
\begin{enumerate}
\item Continuous variables only.  The pairwise mixed model reduces to the familiar multivariate Gaussian parametrized by the symmetric positive-definite inverse covariance matrix $B=\{\beta_{st}\}$ and mean $\mu=B^{-1}\alpha$,
$$
p(x)\propto \exp\left( -\frac{1}{2}(x-B^{-1} \alpha)^{T} B (x-B^{-1}\alpha)\right).
$$
\item Discrete variables only. The pairwise mixed model reduces to a pairwise discrete (second-order interaction) Markov random field,
\begin{equation*}
p(y)\propto\exp{\left(\sum_{j=1}^q \sum_{r=1}^q \phi_{rj}(y_r , y_j)\right)}.
\label{eq:discrete}
\end{equation*}
\end{enumerate}

Although these are the most important aspects, we can characterize the joint distribution further.
The conditional distribution of the continuous variables given the discrete follow a multivariate Gaussian distribution, $p(x|y)= \No(\mu(y),B^{-1})$. Each of these Gaussian distributions share the same inverse covariance matrix $B$ but differ in the mean parameter, since all the parameters are pairwise. By standard multivariate Gaussian calculations,
\begin{align}
p(x|y)&=\No(B^{-1} \gamma(y),B^{-1})\\
\{\gamma(y)\}_s&= \alpha_s+\sum_{j} \rho_{sj}(y_j)\\
p(y)  &\propto \exp{\left(\sum_{j=1}^q \sum_{r=1}^j \phi_{rj} (y_r, y_j) +\frac{1}{2}  \gamma(y)^{T} B^{-1} \gamma(y)\right)}
\end{align}
Thus we see that the continuous variables conditioned on the discrete
are multivariate Gaussian with common covariance, but with means that
depend on the value of the discrete variables. The means depend
additively on the values of the discrete variables since
$\{\gamma(y)\}_s= \sum_{j=1}^r \rho_{sj}(y_j)$. The marginal $p(y)$
has a known form, so for models with few number of discrete variables
we can sample efficiently.

\subsection{Related work on mixed graphical models}
\citet{Lauritzen1996} proposed a type of mixed graphical model, with
the property that conditioned on discrete variables, $p(x|y) =
\No(\mu(y), \Sigma(y) )$. The homogeneous mixed graphical model
enforces common covariance, $\Sigma(y) \equiv \Sigma$. Thus our
proposed model is a special case of Lauritzen's mixed model with the
following assumptions: common covariance, additive mean assumptions
and the marginal $p(y)$ factorizes as a pairwise discrete Markov
random field. With these three assumptions, the full model simplifies
to the mixed pairwise model presented. Although the full model is more
general, the number of parameters scales exponentially with the
number of discrete variables, and the conditional distributions are
not as convenient.  For each state of the discrete variables there is
a mean and covariance. Consider an example with $q$ binary variables
and $p$ continuous variables; the full model requires estimates of
$2^q$ mean vectors and covariance matrices in $p$ dimensions.  Even if
the homogeneous constraint is imposed on Lauritzen's model, there are
still $2^q$ mean vectors for the case of binary discrete
variables. The full mixed model is very complex and cannot be easily
estimated from data without some additional assumptions. In
comparison, the mixed pairwise model has number of parameters
$O((p+q)^2)$ and allows for a natural regularization scheme which
makes it appropriate for high dimensional data. 

An alternative to the regularization approach that we take in this paper, is the limited-order correlation hypothesis testing method \cite{tur2012learning}. The authors develop a hypothesis test via likelihood ratios for conditional independence. However, they restrict to the case where the discrete variables are marginally independent so the maximum likelihood estimates are well-defined for $p>n$.

There is a line of work regarding parameter estimation in undirected
mixed models that are decomposable: any path between two discrete
variables cannot contain only continuous variables. These models allow
for fast exact maximum likelihood estimation through node-wise
regressions, but are only applicable when the structure is known and
$n>p$ \citep{edwards2000introduction}. There is also related work on
parameter learning in directed mixed graphical models. Since our
primary goal is to learn the graph structure, we forgo exact parameter
estimation and use the pseudolikelihood. Similar to the exact maximum
likelihood in decomposable models, the pseudolikelihood can be
interpreted as node-wise regressions that enforce symmetry.

To our knowledge, this work is the first to consider convex optimization procedures for learning the edge structure in mixed graphical models.
\section{Parameter Estimation: Maximum Likelihood and Pseudolikelihood}
\label{sec:paramest}
Given samples $(x_i,y_i)_{i=1}^n$, we want to find the maximum
likelihood estimate of $\Theta$. This can be done by minimizing the
negative log-likelihood of the samples:
\begin{align}
\ell(\Theta)&= -\sum_{i=1}^n \log{ p(x_i,y_i;\Theta)} \mbox{ where }\\
\log{p(x,y;\Theta)}&= \sum_{s=1}^{p}\sum_{t=1}^p -\frac{1}{2} \beta_{st} x_{s} x_{t}+\sum_{s=1}^{p}\alpha_{s} x_{s} +\sum_{s=1}^p \sum_{j=1}^q\rho_{sj}(y_{j})x_{s} \nonumber \\
& +\sum_{j=1}^q\sum_{r=1}^j \phi_{rj}(y_r , y_j)-\log{Z(\Theta)}
\end{align}
The negative log-likelihood is convex, so standard gradient-descent
algorithms can be used for computing the maximum likelihood
estimates. The major obstacle here is $Z(\Theta)$, which involves a high-dimensional integral. Since the pairwise mixed model includes both the discrete
and continuous models as special cases, maximum likelihood estimation
is at least as difficult as the two special cases, the first of  which is a
well-known computationally intractable problem. We defer the
discussion of maximum likelihood estimation to Appendix~\ref{app:mle}.
\subsection{Pseudolikelihood}
\label{sec:pseudolikelihood}
The pseudolikelihood method \cite{besag1975} is a computationally efficient and consistent estimator formed by products of all the conditional distributions:
\begin{align}
\tilde{\ell}(\Theta|x,y)=-\sum_{s=1}^{p} \log{p(x_s|x_{\bs s},y;\Theta)}-\sum_{r=1}^{q} \log{p(y_{r}|x,y_{\bs r};\Theta)} 
\label{eq:negpl}
\end{align}  
The conditional distributions $p(x_{s}| x_{\backslash s}, y; \theta)$ and $p(y_r = k | y_{\backslash r,}, x;\theta)$ take on the familiar form of linear Gaussian and (multiclass) logistic regression, as we pointed out in (\ref{eq:simple1}) and (\ref{eq:simple2}). Here are the details:
\begin{itemize}
\item The conditional distribution of a continuous variable $x_s$ is Gaussian with a linear regression model for the mean, and unknown variance.
\begin{equation}
p(x_s | x_{\backslash s}, y;\Theta)=\frac{\sqrt{\beta_{ss}}}{{\sqrt{2 \pi}}}\exp{ \left(\frac{-\beta_{ss}}{2} \left(\frac{\alpha_s + \sum_{j} \rho_{sj} (y_j) - \sum_{t \neq s} \beta_{st} x_{t} }{\beta_{ss}} -x_{s}\right)^2\right)}
\label{eq:ctscond}
\end{equation}
\item The conditional distribution of a discrete variable $y_r$ with $L_r$ states is a multinomial distribution, as used in (multiclass) logistic regression. Whenever a discrete variable is a predictor, each of its  levels contribute an additive effect; continuous variables contribute linear effects.
\begin{equation}
p(y_r| y_{\backslash r,}, x;\Theta) =\frac{\exp{\left(\sum_{s} \rho_{sr}(y_r) x_{s} +\phi_{rr} (y_r,y_r) +\sum_{j\neq r} \phi_{rj}(y_r,y_{j}) \right)}}{\sum_{l=1}^{L_r } \exp{\left(\sum_{s} \rho_{sr}(l) x_{s} +\phi_{rr} (l,l) +\sum_{j\neq r} \phi_{rj}(l,y_{j}) \right)} } \label{eq:discond}
\end{equation}
\end{itemize}
Taking the negative log of both gives us
\begin{align}
-\log{p(x_s | x_{\backslash s}, y;\Theta)}&= -\frac{1}{2} \log{\beta_{ss}} +\frac{\beta_{ss}}{2} \left(\frac{\alpha_s}{\beta_{ss}}+\sum_{j}\frac{\rho_{sj}(y_j)}{\beta_{ss}} - \sum_{t\neq s} \frac{\beta_{st}}{\beta_{ss}} x_{t} - x_{s}\right)^2\\
-\log{p(y_r| y_{\backslash r,}, x;\Theta)}&=-\log{\frac{\exp{\left(\sum_{s} \rho_{sr}(y_r) x_{s} +\phi_{rr} (y_r,y_r) +\sum_{j\neq r} \phi_{rj}(y_r,y_{j}) \right)}}{\sum_{l=1}^{L_r } \exp{\left(\sum_{s} \rho_{sr}(l) x_{s} +\phi_{rr} (l,l) +\sum_{j\neq r} \phi_{rj}(l,y_{j}) \right)} }}
\end{align}
A generic parameter block, $\theta_{uv}$, corresponding to an edge $(u,v)$ appears twice in the pseudolikelihood, once for each  of the conditional distributions $p(z_u|z_v)$ and $p(z_v|z_u)$. 
\begin{proposition}
The negative log pseudolikelihood in \eqref{eq:negpl} is jointly convex in all the parameters $\{\beta_{ss},\beta_{st}, \alpha_{s}, \phi_{rj}, \rho_{sj}\}$ over the region $\beta_{ss}>0$. 
\label{prop:cvx}
\end{proposition}
We prove Proposition~\ref{prop:cvx} in Appendix~\ref{app:Proofs}.
\subsection{Separate node-wise regression}
A simple approach to parameter estimation is via separate node-wise
regressions; a generalized linear model is used to estimate
$p(z_s|z_{\bs s})$ for each $s$. Separate regressions were used in
\cite{meinshausen06} for the Gaussian graphical model and
\cite{ravikumar2010} for the Ising model.  The method can be thought of as
an asymmetric form of the pseudolikelihood since the pseudolikelihood
enforces that the parameters are shared across the conditionals. Thus
the number of parameters estimated in the separate regression is
approximately double that of the pseudolikelihood, so we expect that
the pseudolikelihood outperforms at low sample sizes and low
regularization regimes. The node-wise regression was used as our
baseline method since it is straightforward to extend it to the mixed
model. As we predicted, the pseudolikelihood or joint procedure
outperforms separate regressions; see top left box of Figures
\ref{fig:sepvspln100} and
\ref{fig:sepvspln10000}. \cite{liu2012distributed,liu2011learning}
confirm that the separate regressions are outperformed by
pseudolikelihood in numerous synthetic settings.

Concurrent work of
\cite{yang2012graphical,yang2013graphical} extend the separate
node-wise regression model from the special cases of Gaussian and
categorical regressions to generalized linear models, where the
univariate conditional distribution of each node $p(x_s|x_{\backslash
  s})$ is specified by a generalized linear model (e.g. Poisson,
categorical, Gaussian). By specifying the conditional distributions,
\cite{besag1974spatial} show that the joint distribution is also
specified. Thus another way to justify our mixed model is to define
the conditionals of a continuous variable as Gaussian linear
regression and the conditionals of a categorical variable as multiple
logistic regression and use the results in \cite{besag1974spatial} to
arrive at the joint distribution in \eqref{eq:jointdensity}.
However, the neighborhood selection algorithm in
\cite{yang2012graphical,yang2013graphical} is restricted to models of
the form $ p(x) \propto \exp\left( \sum_s \theta_s x_s +\sum_{s,t}
  \theta_{st} x_s x_t +\sum_s C(x_s) \right).
$
In particular, this procedure cannot be applied to edge selection in
our pairwise mixed model in \eqref{eq:jointdensity} or the
categorical model in  \eqref{eq:discrete} with greater than 2
states. Our baseline method of separate regressions is closely related
to the neighborhood selection algorithm they proposed; the baseline
can be considered as a generalization of
\cite{yang2012graphical,yang2013graphical} to allow for more general
pairwise interactions with the appropriate regularization to select
edges. Unfortunately, the theoretical results in
\cite{yang2012graphical, yang2013graphical} do not apply to the
baseline nodewise regression method, nor the joint pseudolikelihood.

\section{Conditional Independence and Penalty Terms}
\label{sec:penalty}
In this section, we show how to incorporate edge selection into the maximum likelihood or pseudolikelihood procedures. In the graphical representation of probability distributions, the absence of an edge $e=(u,v)$ corresponds to a conditional independency statement that variables $x_u$ and $x_v$ are conditionally independent given all other variables \citep{koller2009}. We would like to maximize the likelihood subject to a penalization on the number of edges since this results in a sparse graphical model. In the pairwise mixed model, there are 3 type of edges
\begin{enumerate}
\item $\beta_{st}$ is a scalar that corresponds to an edge from $x_s$
  to $x_t$. $\beta_{st}=0$ implies $x_s$ and $x_t$ are conditionally
  independent given all other variables. This parameter is in two
  conditional distributions, corresponding to either $x_s$ or $x_t$ is
  the response variable, $p(x_{s} |x_{\backslash s}, y;\Theta)$ and
  $p(x_{t} |x_{\backslash t}, y;\Theta)$.
\item $\rho_{sj}$ is a vector of length $L_j$. If $\rho_{sj}(y_j) =0 $
  for all values of $y_j$, then $y_j$ and $x_s$ are conditionally
  independent given all other variables. This parameter is in two
  conditional distributions, corresponding to either $x_s$ or $y_j$ being
  the response variable: $p(x_{s} |x_{\backslash s}, y;\Theta)$ and
  $p(y_{j} |x, y_{\backslash j};\Theta)$.
\item $\phi_{rj}$ is a matrix of size $L_r \times L_j$. If $\phi_{rj}
  (y_r, y_j) =0 $ for all values of $y_r$ and $y_j$, then $y_r$ and
  $y_j$ are conditionally independent given all other variables. This
  parameter is in two conditional distributions, corresponding to
  either $y_r$ or $y_j$ being the response variable, $p(y_{r} |x,
  y_{\backslash r};\Theta)$ and $p(y_{j} |x, y_{\backslash
    j};\Theta)$.
\end{enumerate}
For edges that involve discrete variables, the absence of that edge
requires that the entire matrix $\phi_{rj}$ or vector $\rho_{sj}$ is
$0$. The form of the pairwise mixed model motivates the following
regularized optimization problem
\begin{align}
\minimize_{\Theta}~ \ell_{\lambda}(\Theta)=\ell(\Theta)+\lambda\left(\sum_{s<t} \indicator{\beta_{st}\not = 0}  +\sum_{sj} \indicator{\rho_{sj} \not \equiv 0} + \sum_{r<j} \indicator{\phi_{rj} \not \equiv 0} \right)
\end{align}
All parameters that correspond to the same edge are grouped in the same indicator function. This problem is non-convex, so we replace the $l_{0}$ sparsity and group sparsity penalties with the appropriate convex relaxations. For scalars, we use the absolute value ($l_1$ norm), for vectors we use the $l_2$ norm, and for matrices we use the Frobenius norm. This choice corresponds to the standard relaxation from group $l_0$ to group $l_1/l_2$ (group lasso) norm \citep{bach2011optimization,yuan2006model}.
\begin{align}
\minimize_{\Theta}\ \ell_{\lambda}(\Theta)=\ell(\Theta)+ \lambda \left(\sum_{s=1}^p \sum_{t=1}^{s-1} |\beta_{st}| +\sum_{s=1}^p \sum_{j=1}^q \norm{\rho_{sj}}_2 +\sum_{j=1}^q \sum_{r=1}^{j-1} \norm{\phi_{rj}}_F \right)
\label{eq:penpl}
\end{align}
\begin{figure}
\centering
\includegraphics[width=.4\textwidth]{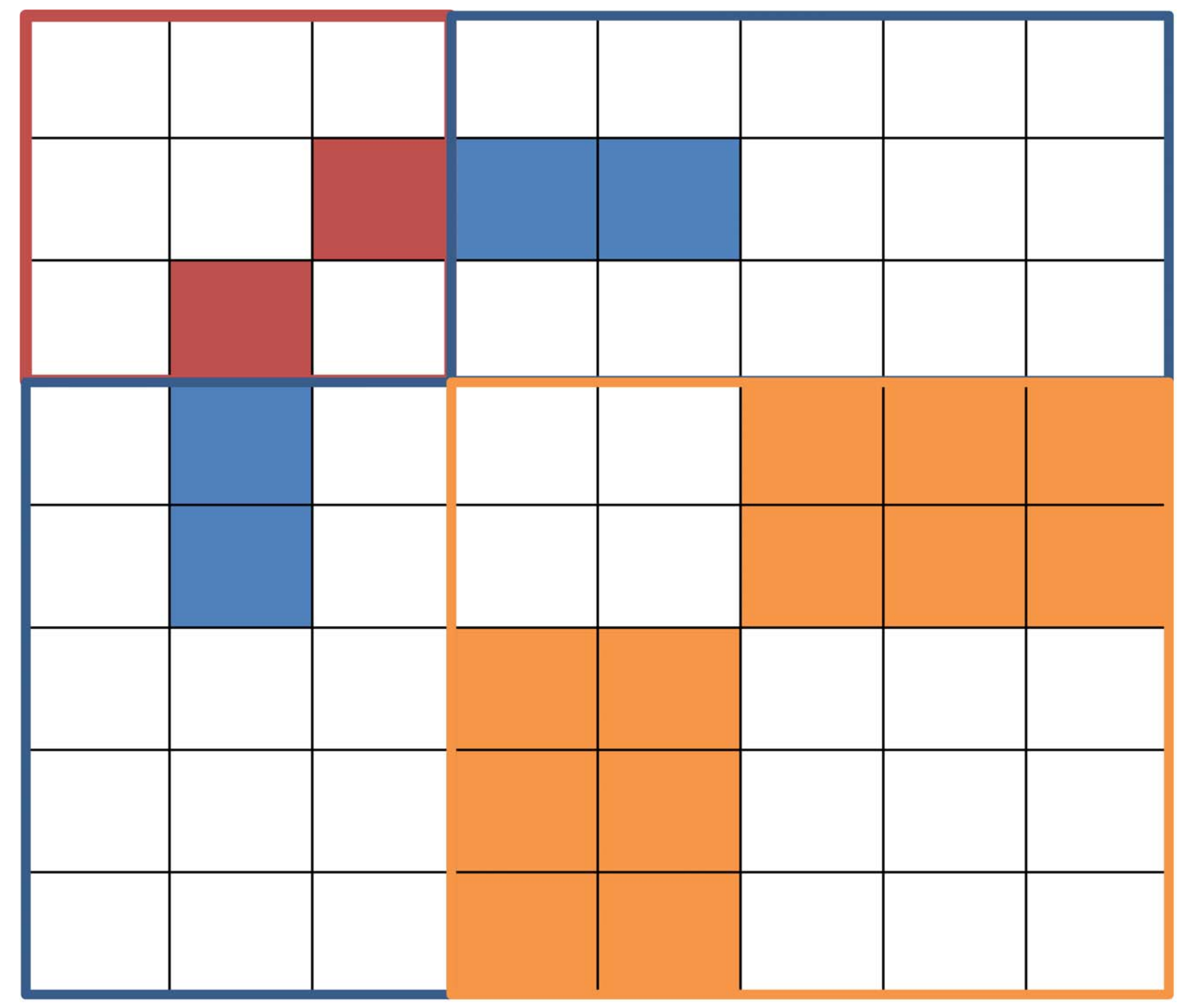}
\caption{\small\em Symmetric matrix represents the parameters $\Theta$ of the model. This example has $p=3$, $q=2$, $L_1 =2$ and $L_2 =3$. The red square corresponds to the continuous graphical model coefficients $B$ and the solid red square is the scalar $\beta_{st}$. The blue square corresponds to the coefficients $\rho_{sj}$ and the solid blue square is a vector of parameters $\rho_{sj} (\cdot)$. The orange square corresponds to the coefficients $\phi_{rj}$ and the solid orange square is a matrix of parameters $\phi_{rj} (\cdot, \cdot)$. The matrix is symmetric, so each parameter block appears in two of the conditional probability regressions.}
\end{figure}

\section{Calibrated regularizers}
\label{sec:calibration}
In  \eqref{eq:penpl} each of the group penalties are treated as equals, irrespective of the size of the group. We suggest a calibration or weighting scheme to balance the load in a more equitable way. We introduce weights for each group of parameters and show how to choose the weights such that each parameter set is treated equally under $p_F$, the fully-factorized independence model \footnote{Under the independence model $p_F$ is fully-factorized $ p(x,y) = \prod_{s=1}^{p} p(x_{s}) \prod_{r=1} ^{q} p(y_{r})$}
\begin{align}
\minimize_{\Theta}\ \ell(\Theta)+ \lambda \left(\sum_{t=1}^p \sum_{t=1}^{s-1} w_{st}|\beta_{st}| +\sum_{s=1}^p \sum_{j=1}^q w_{sj} \norm{\rho_{sj}}_2 +\sum_{j=1}^q \sum_{r=1}^{j-1} w_{rj}\norm{\phi_{rj}}_F \right)
\label{eq:weightpenpl}
\end{align}
Based on the KKT conditions \citep{friedman2007pathwise}, the parameter group $\theta_g$ is non-zero if 
\begin{gather*}
\norm{\frac{\partial \ell}{\partial \theta_{g}}} > \lambda w_{g}
\end{gather*}
where $\theta_{g}$ and $w_{g}$ represents one of the parameter groups and its corresponding weight.
Now $\frac{\partial \ell}{\partial \theta_{g}}$ can be viewed as a generalized residual, and for different groups these are different dimensions---e.g. scalar/vector/matrix. So even under the independence model (when all terms should be zero), one might expect some terms $\norm{\frac{\partial \ell}{\partial \theta_{g}}}$ to have a better random chance of being non-zero  (for example, those of bigger dimensions). 
Thus for all parameters to be on equal footing, we would like to choose the weights $w$ such that
\begin{equation*} 
 E_{p_F}\norm{\frac{\partial \ell}{\partial \theta_{g}}}=\text{constant}\times w_{g}
 \end{equation*}
However, it is simpler to compute in closed form $E_{p_F}\norm{\frac{\partial \ell}{\partial \theta_{g}}}^2$, so we choose $$w_{g} \propto \sqrt{E_{p_F}\norm{\frac{\partial \ell}{\partial \theta_{g}}}^2}$$ where $p_F$ is the fully factorized (independence) model.
In Appendix \ref{app:weights}, we show that the weights can be chosen as 
\begin{align*}
w_{st}&=\sigma_{s} \sigma_{t}\\
w_{sj}&=\sigma_{s} \sqrt{ \sum_{a} p_{a} (1-p_{a})}\\
w_{rj}&=\sqrt{ \sum_{a} p_{a} (1-p_{a}) \sum_{b} q_{b} (1-q_{b})}
\end{align*}
$\sigma_{s}$ is the standard deviation of the continuous variable $x_{s}$. $p_{a} = Pr(y_{r}=a)$ and $q_b= Pr(y_j =b)$ . For all $3$ types of parameters, the weight has the form of $w_{uv} = \mathbf{tr}(\mathbf{cov}(z_{u})) \mathbf{tr} (\mathbf{cov}(z_{v}))$, where $z$ represents a generic variable and $\mathbf{cov}(z)$ is the variance-covariance matrix of $z$.

\section{Model Selection Consistency}
\label{sec:MSC}
In this section, we study the model selection consistency, the correct edge set is selected and the parameter estimates are close to the truth, of pseudolikelihood and maximum likelihood. We will see that the consistency can be established using the framework first developed in \cite{ravikumar2010} and later extended to general m-estimators by \cite{lee2013model}. The proofs in this section are omitted since they follow from a straightforward application of the results in \cite{lee2013model}; the results are stated for the mixed model to show that under certain conditions the estimation procedures are model selection consistent. We also only consider the uncalibrated regularizers to simplify the notation, but it is straightforward to adapt to the calibrated regularizer case.

First, we define some notation. Recall that $\Theta$ is the vector of parameters being estimated $\{\beta_{ss},\beta_{st}, \alpha_{s}, \phi_{rj}, \rho_{sj}\}$, $\Theta^\star$ be the true parameters that estimated the model, and $Q= \nabla^2 \ell (\Theta^\star)$. Both estimation procedures can be written as a convex optimization problem of the form 
\begin{align}
\minimize\ \ell(\Theta) + \lambda \sum_{g \in G} \norm{\Theta_g }_2
\label{eq:generic_estimator}
\end{align}
where $\ell(\theta) = \{ \ell_{ML}, \ell_{PL}\}$ is one of the two log-likelihoods. The regularizer $$
\sum_{g \in G} \norm{\Theta_g} = \lambda \left(\sum_{s=1}^p \sum_{t=1}^{s-1} |\beta_{st}| +\sum_{s=1}^p \sum_{j=1}^q \norm{\rho_{sj}}_2 +\sum_{j=1}^q \sum_{r=1}^{j-1} \norm{\phi_{rj}}_F \right).
$$ The set $G$ indexes the edges $\beta_{st}$, $\rho_{sj}$, and $\phi_{rj}$, and $\Theta_g$ is one of the three types of edges.

It is difficult to establish consistency results for the problem in Equation \eqref{eq:generic_estimator} because the parameters are non-identifiable. This is because $\ell(\Theta)$ is constant with respect to the change of variables $\rho'_{sj} (y_j )=\rho_{sj} (y_j) +c$ and similarly for $\phi$, so we cannot hope to recover $\Theta^\star$. A popular fix for this issue is to drop the last level of $\rho$ and $\phi$, so they are only indicators over $L-1$ levels instead of $L$ levels. This allows for the model to be identifiable, but it results in an asymmetric formulation that treats the last level differently from other levels. Instead, we will maintain the symmetric formulation by introducing constraints. Consider the problem
\begin{equation}
\begin{aligned}
\minimize_{\Theta}\,\ \ell(\Theta) + \lambda \sum_{g \in G} \norm{\Theta_g }_2\\
\text{subject to } C\Theta=0. 
\label{eq:generic_estimator_constrained}
\end{aligned}
\end{equation}
The matrix $C$ constrains the optimization variables such that
\begin{align*}
\sum_{y_j} \rho_{sj}(y_j) = 0 \\
\sum_{y_j } \phi_{rj} ( y_r,y_j ) =0.
\end{align*}
The group regularizer implicitly enforces the same set of constraints, so the optimization problems of Equation \eqref{eq:generic_estimator_constrained} and Equation \eqref{eq:generic_estimator} have the same solutions. For our theoretical results, we will use the constrained formulation of Equation \eqref{eq:generic_estimator_constrained}, since it is identifiable.

We first state some definitions and two assumptions from \cite{lee2013model} that are necessary to present the model selection consistency results. Let $A$ and $I$ represent the active and inactive groups in $\Theta$, so $\Theta^\star _g \neq 0 $ for any $g \in A$ and $\Theta_g ^\star =0$ for any $g \in I$. The sets associated with the active and inactive groups are defined as
\begin{align*}
\mathcal{A} &= \{\Theta\in\mathbb{R}^d : \max_{g\in G}\norm{\Theta_g}_2 \le 1\;\textnormal{and }\norm{\Theta_g}_2 = 0,\,g\in I\} \\
\mathcal{I} &= \{\Theta\in\mathbb{R}^d : \max_{g\in G}\norm{\Theta_g}_2 \le 1\;\textnormal{and }\norm{\Theta_g}_2 = 0,\,g\in A\}.
\end{align*}
Let $M= span(\mathcal{I})^\perp \cap Null(C)$ and $P_M$ be the orthogonal projector onto the subspace $M$. The two assumptions are
\begin{enumerate}
\item Restricted Strong Convexity. We assume that 
\begin{align}
\sup_{v\in M }\frac{v^T \nabla^2 \ell(\Theta) v }{v^T v} \ge m
\label{eq:rsc}
\end{align}
for all $\norm{\Theta-\Theta^\star}_2 \le r$. Since $\nabla^2 \ell(\Theta)$ is lipschitz continuous, the existence of a constant $m$ that satisfies \eqref{eq:rsc} is implied by the pointwise restricted convexity  
$$
\sup_{v\in M }\frac{v^T \nabla^2 \ell(\Theta^\star) v }{v^T v} \ge \tilde{m}.
$$
For convenience, we will use the former.
\item Irrepresentable condition. There exist $\tau\in (0,1)$ such that 
\begin{align}
\sup_{z \in \mathcal{A}}\,V(P_{M^\perp}(\nabla^2 \ell (\Theta^\star)P_M(P_M \nabla^2 \ell (\Theta^\star)P_M)^\dagger P_M z - z)) < 1-\tau,
\label{eq:irrepresentable-condition}
\end{align}
where $V$ is the \emph{infimal convolution} of $\rho_I$, the gauge of set $\mathcal{I}$, and $\mathbbm{1}_{Null(C)^\perp}$:
$$
V(z) = \inf_{z=u_1+u_2}\,\{\rho_I(u_1) + \mathbbm{1}_{Null(C)^\perp}(u_2)\}.
$$
\end{enumerate}
Restricted strong convexity  is a standard assumption that ensures the parameter $\Theta$ is uniquely determined by the value of the likelihood function. Without this, there is no hope of accurately estimating $\Theta^\star$. It is only stated over a subspace $M$ which can be much smaller than $\mathbb{R}^d$. The Irrepresentable condition is a more stringent condition. Intuitively, it requires that the active variables not be overly dependent on the inactive variables. Although the exact form of the condition is not enlightening, it is known to be "almost" necessary for model selection consistency in the lasso \citep{zhao2006model} and a common assumption in other works that establish model selection consistency \citep{ravikumar2010,jalali2011,peng2009}.
We also define the constants that appear in the theorem:
\begin{enumerate}
\item Lipschitz constants $L_1$ and $L_2$. Let $\Lambda(\Theta)$ be the log-partition function. $\Lambda(\Theta)$ and $\ell(\Theta)$ are twice continuously differentiable functions, so their gradient and hessian are locally Lipschitz continuous in a ball of radius $r$ around $\Theta^\star$:
\begin{align*}
\norm{\nabla \Lambda(\Theta_1) -\nabla\Lambda(\Theta_2) }_2 \le L_1 \norm{\Theta_1 - \Theta_2}_2,\ \Theta_1, \Theta_2 \in B_r (\Theta^\star)\\
\norm{\nabla^2 \ell(\Theta_1) -\nabla^2\ell(\Theta_2) }_2 \le L_2 \norm{\Theta_1 - \Theta_2}_2,\ \Theta_1, \Theta_2 \in B_r (\Theta^\star)
\end{align*}
\item Let $\bar{\tau}$ satisfy
$$
\sup_{z \in \mathcal{A}\cup \mathcal{I}}\,V(P_{M^\perp}(\nabla^2 \ell (\Theta^\star)P_M(P_M \nabla^2 \ell (\Theta^\star)P_M)^\dagger P_M z - z)) < \bar{\tau}.
$$
$V$ is a continuous function of $z$, so a finite $\bar{\tau}$ exists.
\end{enumerate}

\begin{theorem}
\label{cor:penalized-mle-consistent}
Suppose we are given samples $x^{(1)},\dots,x^{(n)}$ from the mixed model with unknown parameters $\Theta^\star$. If we select 
$$
\lambda = \frac{2\sqrt{256L_1}\bar{\tau}}{\tau}\sqrt{\frac{(\max_{g\in G}|g|)\log|G|}{n}}
$$
and the sample size $n$ is larger than
$$
\max\,\begin{cases} \frac{4096L_1L_2^2\bar{\tau}^2}{m^4\tau^4}\left(2 + \frac{\tau}{\bar{\tau}}\right)^4(\max_{g\in G}|g|)|A|^2\log|G| \\
\frac{2048L_1}{m^2r^2}(2 + \frac{\tau}{\bar{\tau}})^2(\max_{g\in G}|g|)|A|\log|G|,
\end{cases}
$$
then, with probability at least $1-2\big(\max_{g\in G}|g|\big)\exp(-c\lambda^2n)$, the optimal solution to \eqref{eq:generic_estimator} is unique and model selection consistent, 
\begin{enumerate}
\item $\|\hat{\Theta} - \Theta^\star\|_2 \le \frac{4}{m}\left(\frac{\bar{\tau}+1}{2\tau}\right)\sqrt{\frac{256L_1 |A|(\max_{g\in G}|g|)\log|G|}{n}},$
\item $\hat{\Theta}_g = 0,\,g\in I$ and $\hat{\Theta}_g \ne 0\;\text{if }\norm{\Theta^\star_g}_2 > \frac{1}{m}\left(1 + \frac{\tau}{2\bar{\tau}}\right)\sqrt{|A|}\lambda$.
\end{enumerate}

\end{theorem}
\begin{remark}
The same theorem applies to both the maximum likelihood and pseudolikelihood estimators. For the maximum likelihood, the constants can be tightened; everywhere $L_1$ appears can be replaced by $L_1 /128$ and the theorem remains true. However, the values of $\tau, \bar{\tau}, m, L_1,L_2$ are different for the two methods. For the maximum likelihood, the gradient of the log-partition $\nabla\Lambda(\Theta)$ and hessian of the log-likelihood $\nabla^2 \ell(\Theta)$ do not depend on the samples. Thus the constants $\tau, \bar{\tau}, m, L_1,L_2$ are completely determined by $\Theta^\star$ and the likelihood. For the pseudolikelihood, the values of $\tau, \bar{\tau}, m,L_2$ depend on the samples, and the theorem only applies if the assumptions are made on sample quantities; thus, the theorem is less useful in practice when applied to the pseudolikelihood.  This is similar to the situation in \cite{yang2013graphical}, where assumptions are made on sample quantities. 
\end{remark}

\section{Optimization Algorithms}
\label{sec:optalg}
In this section, we discuss two algorithms for solving \eqref{eq:penpl}: the proximal gradient and the proximal newton methods. This is a convex optimization problem that decomposes into the form $f(x)+g(x)$, where $f$ is smooth and convex and $g$ is convex but possibly non-smooth. In our case $f$ is the negative log-likelihood or negative log-pseudolikelihood and $g$ are the group sparsity penalties.

Block coordinate descent is a frequently used method when the non-smooth function $g$ is the $l_1$ or group $l_1$. It is especially easy to apply when the function $f$ is quadratic, since each block coordinate update can be solved in closed form for many different non-smooth $g$ \citep{friedman2007pathwise}. The smooth $f$ in our particular case is not quadratic, so each block update cannot be solved  in closed form. However in certain problems (sparse inverse covariance), the update can be approximately solved by using an appropriate inner optimization routine \citep{glasso}.
\subsection{Proximal Gradient}
Problems of this form are well-suited for the proximal gradient and accelerated proximal gradient algorithms as long as the proximal operator of $g$ can be computed \citep{combettes2011proximal,beck2010gradient}
\begin{align}
prox_{t}(x)=\argmin_{u} \frac{1}{2t}\norm{x-u}^2+g(u)
\end{align}
For the sum of $l_2$ group sparsity penalties considered, the proximal operator takes the familiar form of soft-thresholding and group soft-thresholding \citep{bach2011optimization}. Since the groups are non-overlapping, the proximal operator simplifies to scalar soft-thresholding for $\beta_{st}$ and group soft-thresholding for $\rho_{sj}$ and $\phi_{rj}$.

The class of proximal gradient and accelerated proximal gradient algorithms is directly applicable to our problem. These algorithms work by solving a first-order model at the current iterate $x_{k}$
\begin{align}
\argmin_{u}~ &f(x_{k})+\nabla f(x_{k}) ^{T}(u-x_{k}) +\frac{1}{2t}\norm{u-x_{k}}^2 +g(u)\\
&=\argmin_{u}~ \frac{1}{2t}\norm{u-\left(x_{k}-t\nabla f(x_{k})\right)}^2+g(u)\\
&=prox_{t} (x_{k} -t\nabla f(x_{k}))
\end{align}
The proximal gradient iteration is given by $x_{k+1} = prox_{t} \left( x_{k} - t \nabla f(x_{k}) \right)$ where $t$ is determined by line search. The theoretical convergence rates and properties of the proximal gradient algorithm and its accelerated variants are well-established \citep{beck2010gradient}. The accelerated proximal gradient method achieves linear convergence rate of $O(c^k)$ when the objective is strongly convex and the sublinear rate $O(1/k^2)$ for non-strongly convex problems.

The TFOCS framework \citep{becker2011} is a package that allows us to experiment with $6$ different variants of the accelerated proximal gradient algorithm. The TFOCS authors found that the Auslender-Teboulle algorithm exhibited less oscillatory behavior, and proximal gradient experiments in the next section were done using the Auslender-Teboulle implementation in TFOCS. 
\subsection{ Proximal Newton Algorithms}
\label{sec:proxnewton}
This section borrows heavily from \cite{schmidt2010}, \cite{schmidt2011} and \cite{lee2012proximal}. The class of proximal Newton algorithms is a 2nd order analog of the proximal gradient algorithms with a quadratic convergence rate \citep{lee2012proximal}. It attempts to incorporate 2nd order information about the smooth function $f$ into the model function. At each iteration, it minimizes a quadratic model centered at $x_{k}$
\begin{align}
&\argmin_{u}~ f(x_{k})+\nabla f(x_{k})^{T} (u-x_{k})+\frac{1}{2t}(u-x_{k})^{T}H(u-x_{k}) +g(u)\\
&=\argmin_{u}~ \frac{1}{2t}\left(u-x_{k}+tH^{-1} \nabla f(x_{k}) \right)^{T} H \left(u-x_{k}+tH^{-1} \nabla f(x_{k}) \right)+g(u)\\
&=\argmin_{u}~ \frac{1}{2t} \norm{u-\left( x_{k}-tH^{-1}\nabla f(x_{k}) \right)}^{2}_{H}+g(u)\\
&:=Hprox_{t}\left(x_{k}-tH^{-1}\nabla f(x_{k}) \right)
\mbox{ where } H= \nabla^{2} f(x_{k})
\end{align}
\begin{algorithm}
\caption{Proximal Newton}
\begin{algorithmic}
\Repeat
\State Solve subproblem $p_{k} = Hprox_{t}\left(x_{k}-tH_{k}^{-1}\nabla f(x_{k}) \right) - x_{k}$ using TFOCS.
\State Find $t$ to satisfy Armijo line search condition with parameter $\alpha$
\begin{displaymath}
f(x_{k} + tp_{k}) +g(x_{k} +tp_{k}) \leq f(x_{k})+g(x_{k}) - \frac{t\alpha}{2} \norm{p_{k}}^2
\end{displaymath}
\State Set $x_{k+1}=x_{k}+tp_{k}$
\State $k=k+1$
\Until{$\frac{\norm{x_{k} - x_{k+1}}}{\norm{x_{k} }} < tol$} 
\end{algorithmic}
\end{algorithm}
The $Hprox$ operator is analogous to the proximal operator, but in the $\norm{\cdot}_{H}$-norm. It simplifies to the proximal operator if $H=I$, but in the general case of positive definite $H$ there is no closed-form solution for many common non-smooth $g(x)$ (including $l_{1}$ and group $l_{1}$). However if the proximal operator of $g$ is available, each of these sub-problems can be solved efficiently with proximal gradient. In the case of separable $g$, coordinate descent is also applicable. Fast methods for solving the subproblem $Hprox_{t} (x_k - t H^{-1} \nabla f(x_{k}))$ include coordinate descent methods, proximal gradient methods, or Barzilai-Borwein \citep{friedman2007pathwise,combettes2011proximal,beck2010gradient, wright2009sparse}. The proximal Newton framework allows us to bootstrap many previously developed solvers to the case of arbitrary loss function $f$.

Theoretical analysis in \cite{lee2012proximal} suggests that proximal Newton methods generally require fewer outer iterations (evaluations of $Hprox$) than first-order methods while providing higher accuracy because they incorporate 2nd order information. We have confirmed empirically that the proximal Newton methods are faster when $n$ is very large or the gradient is expensive to compute (e.g. maximum likelihood estimation). Since the objective is quadratic, coordinate descent is also applicable to the subproblems. The hessian matrix $H$ can be replaced by a quasi-newton approximation such as BFGS/L-BFGS/SR1. In our implementation, we use the \texttt{PNOPT} implementation \citep{lee2012proximal}. 
\subsection{Path Algorithm}
Frequently in machine learning and statistics, the regularization parameter $\lambda$ is heavily dependent on the dataset. $\lambda$ is generally chosen via cross-validation or holdout set performance, so it is convenient to provide solutions over an interval of $[\lambda_{min} , \lambda_{max}]$. We start the algorithm at $\lambda_1 = \lambda_{max}$ and solve, using the previous solution as warm start, for $\lambda_2 > \ldots> \lambda_{min}$. We find that this reduces the cost of fitting an entire path of solutions (See Figure \ref{fig:modelselect}). $\lambda_{max}$ can be chosen as the smallest value such that all parameters are $0$ by using the KKT equations \citep{friedman2007pathwise}. 

\section{Conditional Model}
\label{sec:condmodel}
We can generalize our mixed model, when there are additional features $f$, to a class of conditional random
fields. Conditional models only model the conditional distribution
$p(z|f)$, as opposed to the joint distribution $p(z,f)$, where $z$ are
the variables of interest to the prediction task and $f$ are
features. 

In addition to observing $x$ and $y$, we observe features $f$ and we build a graphical model for the conditional distribution $p(x,y|f)$. Consider a full pairwise model $p(x,y,f)$ of the form \eqref{eq:jointdensity}. We then choose to only model the joint distribution over only the variables $x$ and $y$ to give us $p(x,y|f)$ which is of the form
\begin{align}
p(x,y|f;\Theta)=&\frac{1}{Z(\Theta |f)} \exp  \left( \sum_{s=1}^p \sum_{t=1}^{p}-\frac{1}{2} \beta_{st} x_{s} x_{t}+\sum_{s=1}^{p}\alpha_{s} x_{s} +\sum_{s=1}^p \sum_{j=1}^{q} \rho_{sj}(y_{j})x_{s} \right. \nonumber \\ 
& \left.+\sum_{j=1}^q \sum_{r=1}^j \phi_{rj}(y_r , y_j) +\sum_{l=1}^{F} \sum_{s=1}^p \gamma_{ls} x_s f_l + \sum_{l=1}^F \sum_{r=1}^{q} \eta_{lr}(y_r) f_l \right)
\label{eq:jointdensityfeat}
\end{align}
We can also consider a more general model where each pairwise edge potential depends on the features
\begin{align}
p(x,y|f;\Theta)=\frac{1}{Z(\Theta |f)} &\exp\left(\sum_{s=1}^p\sum_{t=1}^{p}-\frac{1}{2} \beta_{st}(f) x_{s} x_{t}+\sum_{s=1}^{p}\alpha_{s}(f) x_{s} \right. \nonumber \\ & \left. +\sum_{s=1}^p\sum_{j=1}^{q}\rho_{sj}(y_{j},f)x_{s}+\sum_{j=1}^{q}\sum_{r=1}^j \phi_{rj}(y_r , y_j,f) \vphantom{\sum_{s=1}^p\sum_{t=1}^{p}} \right)
\end{align}
\eqref{eq:jointdensityfeat} is a special case of this where only the node potentials depend on features and the pairwise potentials are independent of feature values.  The specific parametrized form we consider is $\phi_{rj}(y_r, y_j,f)\equiv \phi_{rj}(y_r,y_j)$ for $r\neq j$, $\rho_{sj}(y_j,f)\equiv \rho_{sj}(y_j)$, and $\beta_{st}(f)=\beta_{st}$. The node potentials depend linearly on the feature values, $\alpha_{s}(f)=\alpha_{s} + \sum_{l=1}^{F} \gamma_{ls} x_s f_l$, and $\phi_{rr}(y_r,y_r,f) = \phi_{rr}(y_r,y_r) + \sum_{l} \eta_{lr}(y_r)$.
\section{Experimental Results}
\label{sec:exp}
We present experimental results on synthetic data, survey data and on a conditional model.
\subsection{Synthetic Experiments}
\begin{figure}
\centering
  \subfloat[][Graph Structure with $4$ continuous and $4$ discrete variables.]{
    \label{fig:syntheticgraph}
        \includegraphics[width=.4\textwidth]{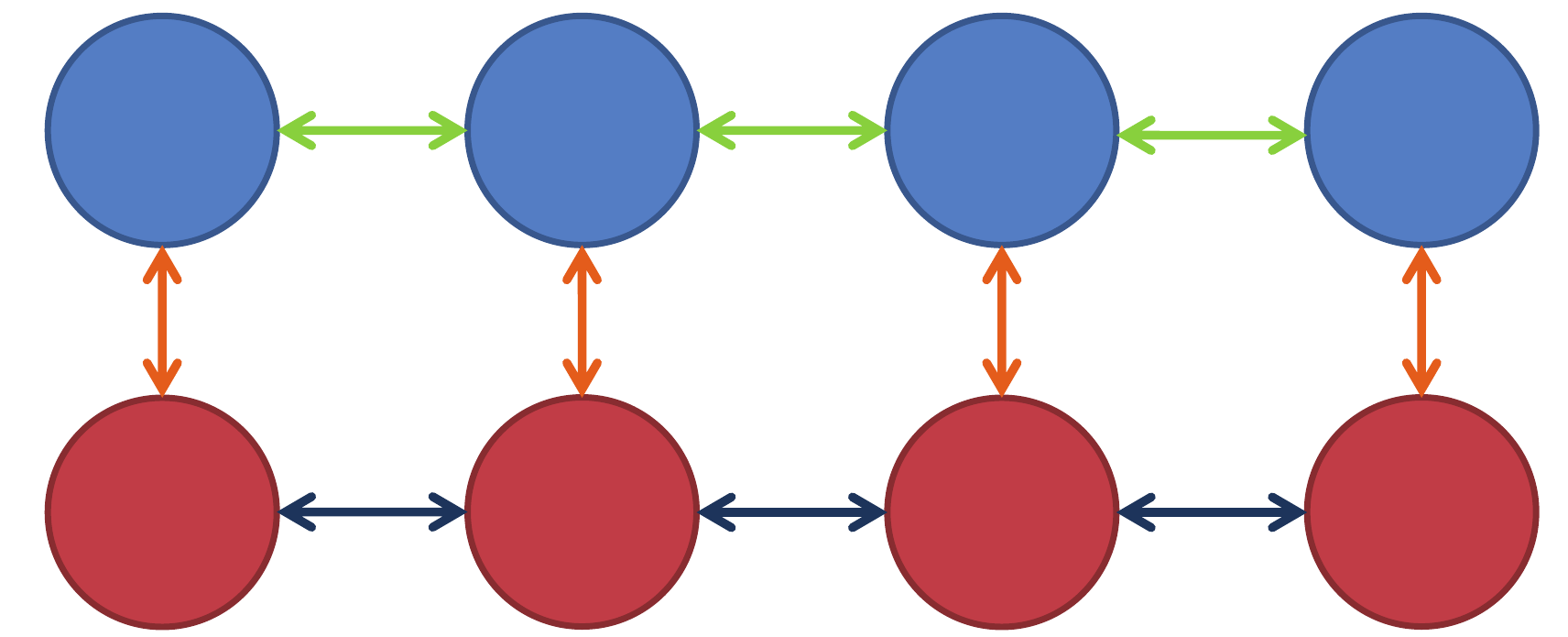}       
  }\hfill
   \subfloat[][Probability of recovery]{
    \label{fig:syntheticplot}
        \includegraphics[width=.4\textwidth]{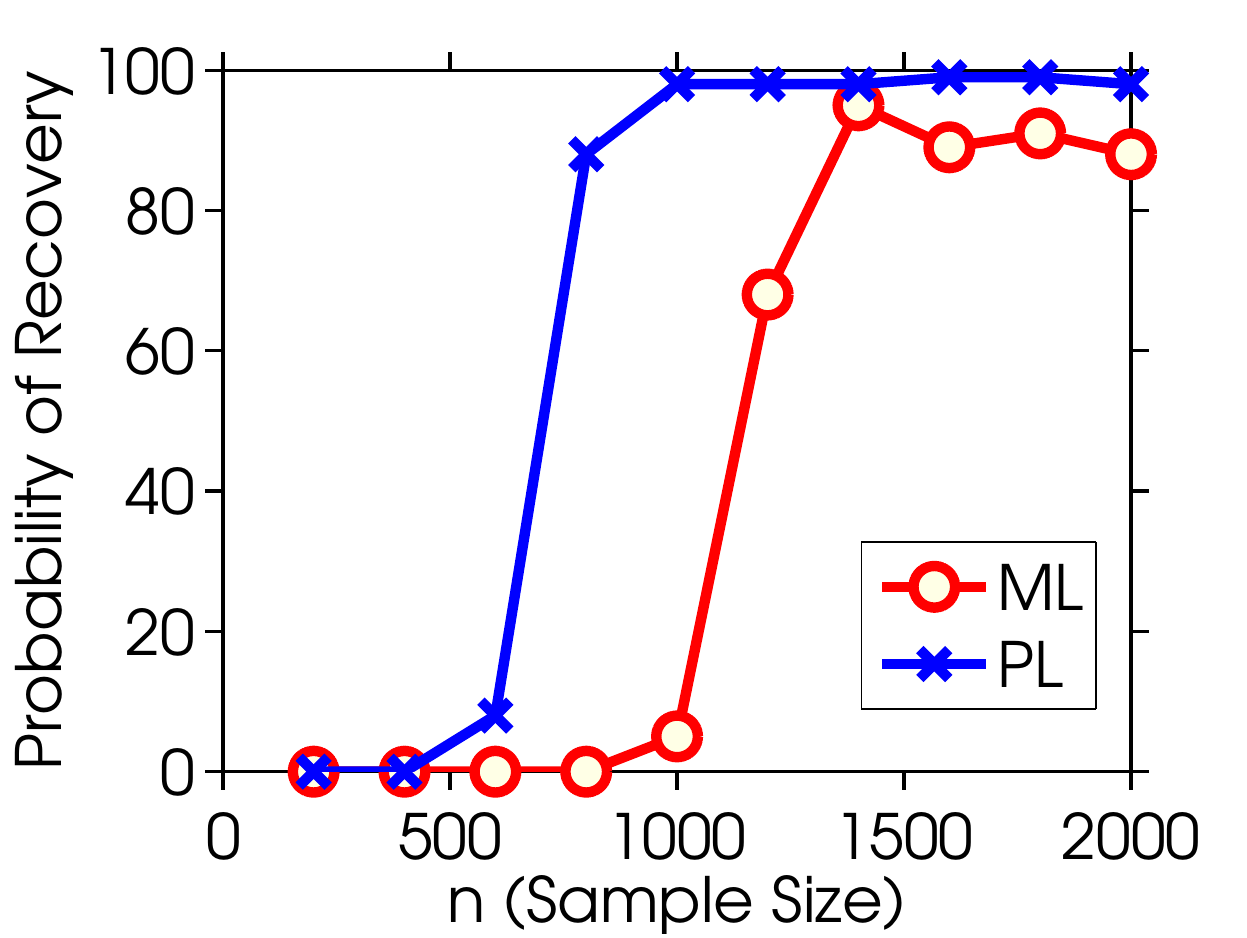}
  }\hfill
 \caption{\small\em Figure \ref{fig:syntheticgraph} shows the graph used in the synthetic experiments for $p=q=4$; the experiment used $p$=10 and $q$=10. Blue nodes are continuous variables, red nodes are binary variables and the orange, green and dark blue lines represent the $3$ types of edges. Figure \ref{fig:syntheticplot} is a plot of the probability of correct edge recovery at a given sample size using Maximum Likelihood and Pseudolikelihood. Results are averaged over $100$ trials.}
\end{figure}

In the synthetic experiment, the training points are sampled from a true model with $10$ continuous variables and $10$ binary variables. The edge structure is shown in Figure \ref{fig:syntheticgraph}. $\lambda$ is chosen as $5
\sqrt{\frac{\log{p+q}}{n}}$ as suggested by the theoretical results in Section \ref{sec:MSC}. We see from the experimental results that recovery of the correct edge set undergoes a sharp phase transition, as expected. With $n=1000$ samples, the pseudolikelihood is recovering the correct edge set with probability nearly $1$. The phase transition experiments were done using the proximal Newton algorithm discussed in Section \ref{sec:proxnewton}.
\subsection{Survey Experiments}
The census survey dataset we consider consists of $11$ variables, of
which $2$ are continuous and $9$ are discrete: age (continuous),
log-wage (continuous), year($7$ states), sex($2$ states),marital
status ($5$ states), race($4$ states), education level ($5$ states),
geographic region($9$ states), job class ($2$ states), health ($2$
states), and health insurance ($2$ states). The dataset was assembled
by Steve Miller of OpenBI.com from the March 2011 Supplement to
Current Population Survey data. All the evaluations are done using a
holdout test set of size $100,000$ for the survey experiments. The
regularization parameter $\lambda$ is varied over the interval
$[5\times 10^{-5}, 0.7]$ at $50$ points equispaced on log-scale for all
experiments.
\subsubsection{Model Selection}
\begin{figure}
\centering
\includegraphics[width=3in]{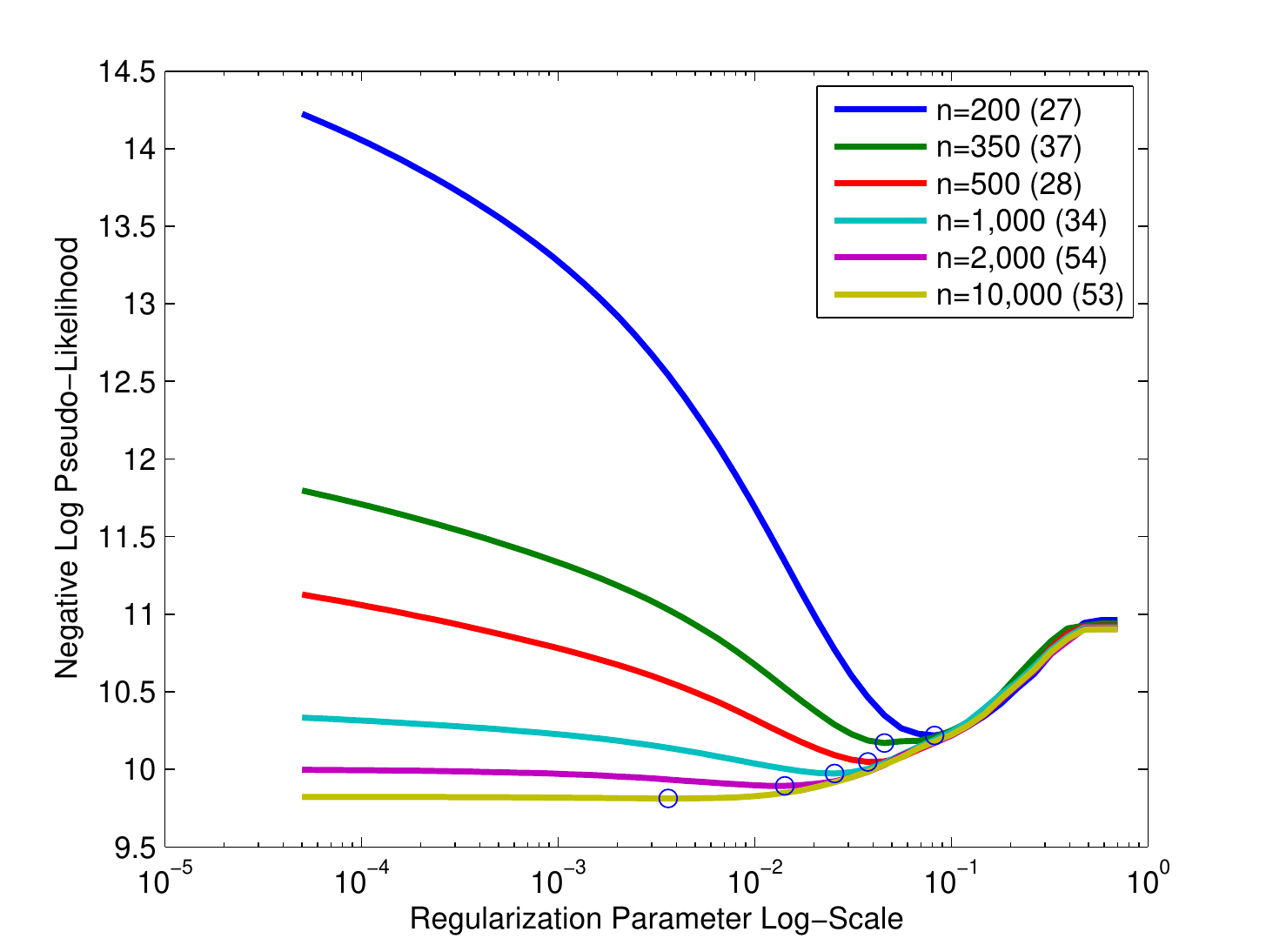}
\caption{\small\em Model selection under different training set sizes. Circle denotes the lowest test set negative log pseudolikelihood and the number in parentheses is the number of edges in that model at the lowest test negative log pseudolikelihood. The saturated model has $55$ edges.}
\label{fig:modelselect}
\end{figure}
In Figure \ref{fig:modelselect}, we study the model selection performance of learning a graphical model over the $11$ variables under different training samples sizes. We see that as the sample size increases, the optimal model is increasingly dense, and less regularization is needed. 

\subsubsection{Comparing against Separate Regressions}
\begin{figure}
\centering
\includegraphics[width=.9\textwidth]{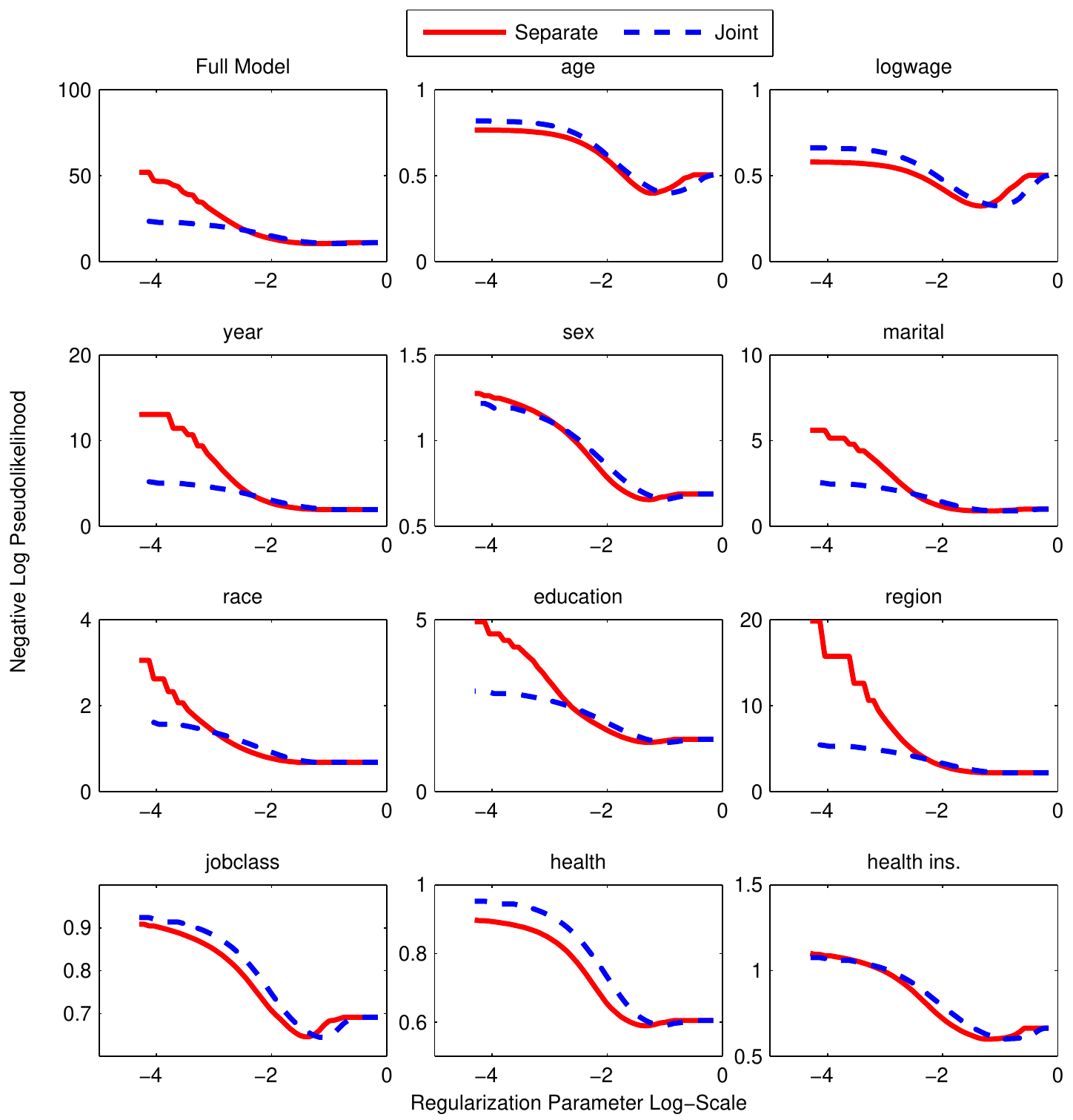}
\caption{\small\em Separate Regression vs Pseudolikelihood $n=100$. $y$-axis is the appropriate regression loss for the response variable. For low levels of regularization and at small training sizes, the pseudolikelihood seems to overfit less; this may be due to a global regularization effect from fitting the joint distribution as opposed to separate regressions. }
\label{fig:sepvspln100}
\end{figure}

\begin{figure}
\centering
\includegraphics[width=.9\textwidth]{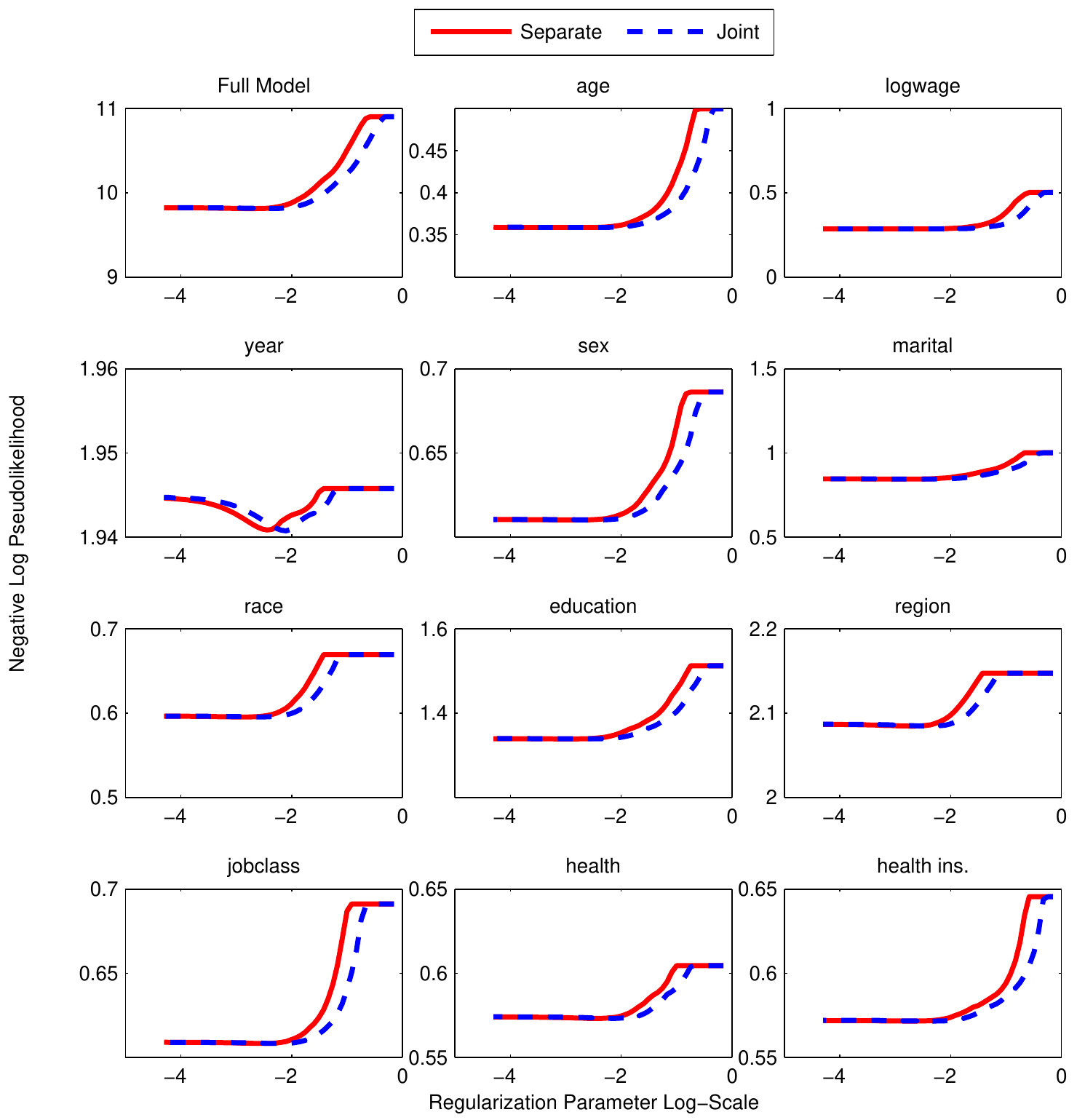}
\caption{\small\em Separate Regression vs Pseudolikelihood $n=10,000$. $y$-axis is the appropriate regression loss for the response variable. At large sample sizes, separate regressions and pseudolikelihood perform very similarly. This is expected since this is nearing the asymptotic regime.}
\label{fig:sepvspln10000}
\end{figure}

A sensible baseline method to compare against is a separate regression algorithm. This algorithm fits a linear Gaussian or (multiclass) logistic regression of each variable conditioned on the rest. We can evaluate the performance of the pseudolikelihood by evaluating $-\log{p(x_{s}|x_{\backslash s},y)}$ for linear regression and $-\log{p(y_{r}|y_{\backslash r},x)}$ for (multiclass) logistic regression. Since regression is directly optimizing this loss function, it is expected to do better. The pseudolikelihood objective is similar, but has half the number of parameters as the separate regressions since the coefficients are shared between two of the conditional likelihoods. From Figures \ref{fig:sepvspln100} and \ref{fig:sepvspln10000}, we can see that the pseudolikelihood performs very similarly to the separate regressions and sometimes even outperforms regression. The benefit of the pseudolikelihood is that we have learned parameters of the joint distribution $p(x,y)$ and not just of the conditionals $p(x_{s}|y,x_{\backslash s})$. On the test dataset, we can compute quantities such as conditionals over arbitrary sets of variables $p(y_{A}, x_{B}|y_{A^{C}},x_{B^C})$ and marginals $p(x_{A},y_{B})$ \citep{koller2009}. This would not be possible using the separate regressions. 

\subsubsection{Conditional Model}
\begin{figure}

\centering
\includegraphics[width=.9\textwidth]{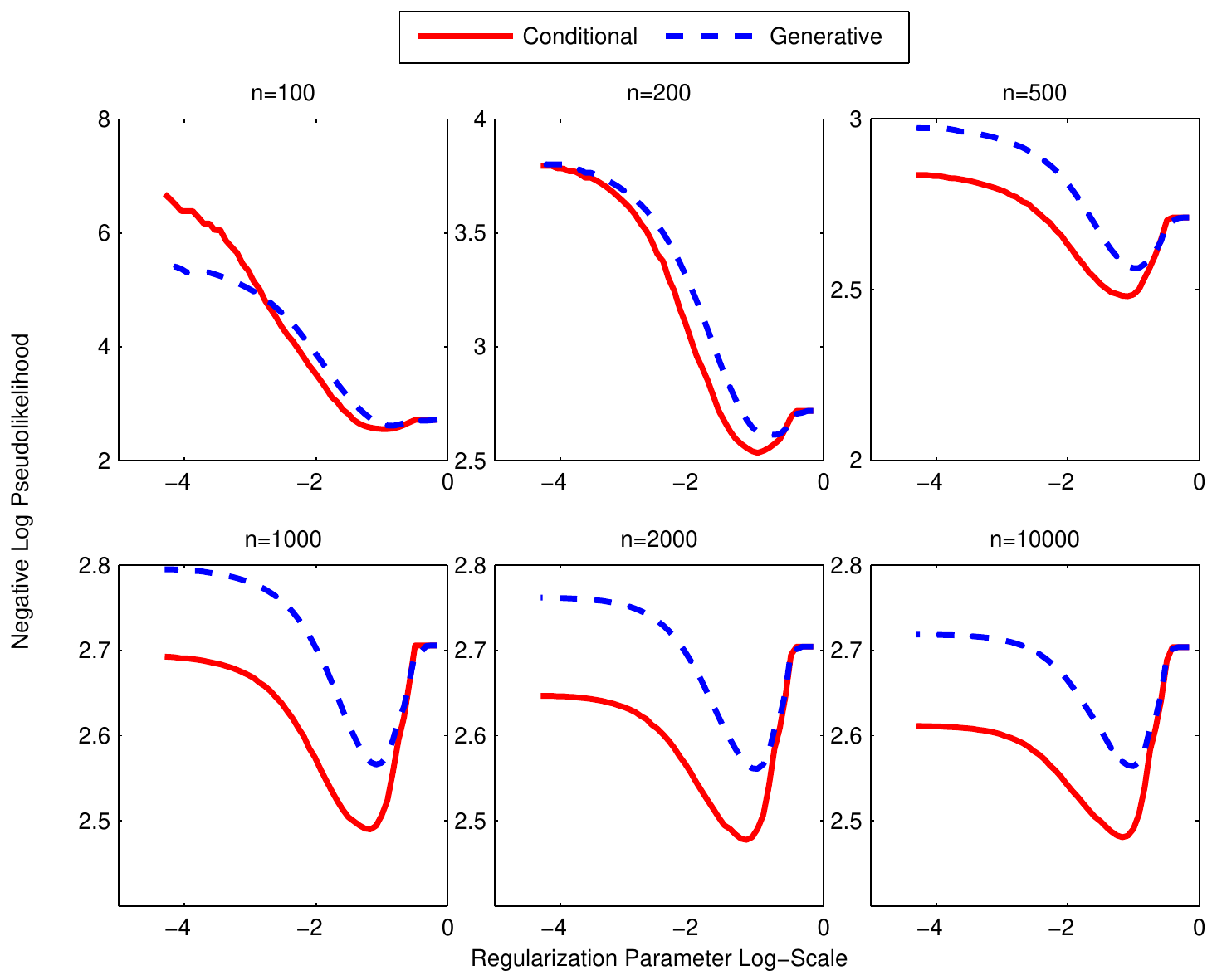}
\caption{\small\em Conditional Model vs Generative Model at various sample sizes. $y$-axis is test set performance is evaluated on negative log pseudolikelihood of the conditional model. The conditional model outperforms the full generative model at except the smallest sample size $n=100$.}
\label{fig:condvsgen}
\end{figure}

Using the conditional model \eqref{eq:jointdensityfeat}, we model only the $3$ variables logwage, education($5$) and jobclass($2$). The other $8$ variables are only used as features. The conditional model is then trained using the pseudolikelihood. We compare against the generative model that learns a joint distribution on all $11$ variables. From Figure \ref{fig:condvsgen}, we see that the conditional model outperforms the generative model, except at small sample sizes. This is expected since the conditional distribution models less variables. At very small sample sizes and small $\lambda$, the generative model outperforms the conditional model. This is likely because generative models converge faster (with less samples) than discriminative models to its optimum. 
\subsubsection{Maximum Likelihood vs Pseudolikelihood}
\begin{figure}
\centering
\includegraphics[width=.9\textwidth]{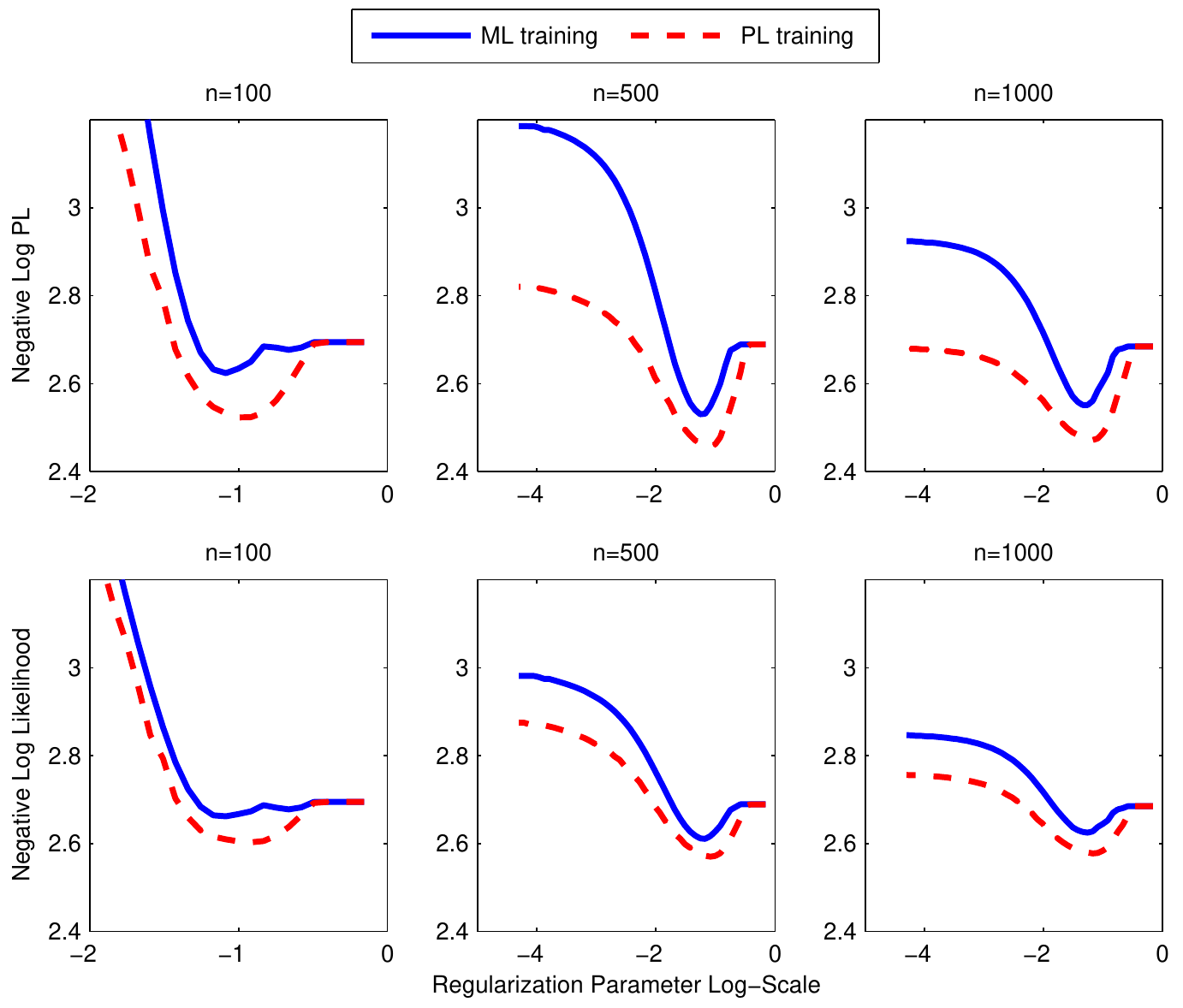}
\caption{\small\em Maximum Likelihood vs Pseudolikelihood. $y$-axis for top row is the negative log pseudolikelihood. $y$-axis for bottom row is the negative log likelihood. Pseudolikelihood outperforms maximum likelihood across all the experiments.}
\label{fig:likvspl}
\end{figure}

The maximum likelihood estimates are computable for very small models such as the conditional model previously studied. The pseudolikelihood was originally motivated as an approximation to the likelihood that is computationally tractable. We compare the maximum likelihood and maximum pseudolikelihood on two different evaluation criteria: the negative log likelihood and negative log pseudolikelihood. In Figure \ref{fig:likvspl}, we find that the pseudolikelihood outperforms maximum likelihood under both the negative log likelihood and negative log pseudolikelihood. We would expect that the pseudolikelihood trained model does better on the pseudolikelihood evaluation and maximum likelihood trained model does better on the likelihood evaluation. However, we found that the pseudolikelihood trained model outperformed the maximum likelihood trained model on both evaluation criteria. Although asymptotic theory suggests that maximum likelihood is more efficient than the pseudolikelihood, this analysis is applicable because of the finite sample regime and misspecified model. See \citet{liang2008asymptotic} for asymptotic analysis of pseudolikelihood and maximum likelihood under a well-specified model. We also observed the pseudolikelihood slightly outperforming the maximum likelihood in the synthetic experiment of Figure \ref{fig:syntheticplot}.

\section*{Acknowledgements}
We would like to thank Percy Liang and Rahul Mazumder for helpful discussions. The work on consistency follows from a collaboration with Yuekai Sun and Jonathan Taylor. Jason Lee is supported by the Department of Defense (DoD) through the National Defense Science \& Engineering Graduate Fellowship (NDSEG) Program, National Science Foundation Graduate Research Fellowship Program, and the Stanford Graduate Fellowship. Trevor Hastie was partially supported by grant DMS-1007719 from the National Science Foundation, and grant RO1-EB001988-15 from the National Institutes of Health.

\bibliographystyle{imsart-nameyear}
\bibliography{struct_graphmodel}

\begin{thebibliography}{39}
\providecommand{\natexlab}[1]{#1}
\providecommand{\url}[1]{\texttt{#1}}
\expandafter\ifx\csname urlstyle\endcsname\relax
  \providecommand{\doi}[1]{doi: #1}\else
  \providecommand{\doi}{doi: \begingroup \urlstyle{rm}\Url}\fi

\bibitem[Bach et~al.(2011)Bach, Jenatton, Mairal, and
  Obozinski]{bach2011optimization}
F.~Bach, R.~Jenatton, J.~Mairal, and G.~Obozinski.
\newblock Optimization with sparsity-inducing penalties.
\newblock \emph{Foundations and Trends in Machine Learning}, 4:\penalty0
  1--106, 2011.
\newblock URL \url{http://dx.doi.org/10.1561/2200000015}.

\bibitem[Banerjee et~al.(2008)Banerjee, El~Ghaoui, and
  d'Aspremont]{banerjee2008}
O.~Banerjee, L.~El~Ghaoui, and A.~d'Aspremont.
\newblock Model selection through sparse maximum likelihood estimation for
  multivariate gaussian or binary data.
\newblock \emph{The Journal of Machine Learning Research}, 9:\penalty0
  485--516, 2008.

\bibitem[Beck and Teboulle(2010)]{beck2010gradient}
A.~Beck and M.~Teboulle.
\newblock Gradient-based algorithms with applications to signal recovery
  problems.
\newblock \emph{Convex Optimization in Signal Processing and Communications},
  pages 42--88, 2010.

\bibitem[Becker et~al.(2011)Becker, Cand{\`e}s, and Grant]{becker2011}
S.R. Becker, E.J. Cand{\`e}s, and M.C. Grant.
\newblock Templates for convex cone problems with applications to sparse signal
  recovery.
\newblock \emph{Mathematical Programming Computation}, pages 1--54, 2011.

\bibitem[Besag(1974)]{besag1974spatial}
J.~Besag.
\newblock Spatial interaction and the statistical analysis of lattice systems.
\newblock \emph{Journal of the Royal Statistical Society. Series B
  (Methodological)}, pages 192--236, 1974.

\bibitem[Besag(1975)]{besag1975}
J.~Besag.
\newblock Statistical analysis of non-lattice data.
\newblock \emph{The statistician}, pages 179--195, 1975.

\bibitem[Combettes and Pesquet(2011)]{combettes2011proximal}
P.L. Combettes and J.C. Pesquet.
\newblock Proximal splitting methods in signal processing.
\newblock \emph{Fixed-Point Algorithms for Inverse Problems in Science and
  Engineering}, pages 185--212, 2011.

\bibitem[Edwards(2000)]{edwards2000introduction}
D.~Edwards.
\newblock \emph{Introduction to graphical modelling}.
\newblock Springer, 2000.

\bibitem[Friedman et~al.(2007)Friedman, Hastie, H{\"o}fling, and
  Tibshirani]{friedman2007pathwise}
J.~Friedman, T.~Hastie, H.~H{\"o}fling, and R.~Tibshirani.
\newblock Pathwise coordinate optimization.
\newblock \emph{The Annals of Applied Statistics}, 1\penalty0 (2):\penalty0
  302--332, 2007.

\bibitem[Friedman et~al.(2008{\natexlab{a}})Friedman, Hastie, and
  Tibshirani]{friedman2008}
J.~Friedman, T.~Hastie, and R.~Tibshirani.
\newblock Sparse inverse covariance estimation with the graphical lasso.
\newblock \emph{Biostatistics}, 9\penalty0 (3):\penalty0 432--441,
  2008{\natexlab{a}}.

\bibitem[Friedman et~al.(2008{\natexlab{b}})Friedman, Hastie, and
  Tibshirani]{glasso}
J.~Friedman, T.~Hastie, and R.~Tibshirani.
\newblock Sparse inverse covariance estimation with the graphical lasso.
\newblock \emph{Biostatistics}, 9\penalty0 (3):\penalty0 432--441,
  2008{\natexlab{b}}.

\bibitem[Friedman et~al.(2010)Friedman, Hastie, and Tibshirani]{friedman2010}
J.~Friedman, T.~Hastie, and R.~Tibshirani.
\newblock Applications of the lasso and grouped lasso to the estimation of
  sparse graphical models.
\newblock Technical report, Technical Report, Stanford University, 2010.

\bibitem[Guo et~al.(2010)Guo, Levina, Michailidis, and Zhu]{guo2010joint}
J.~Guo, E.~Levina, G.~Michailidis, and J.~Zhu.
\newblock Joint structure estimation for categorical markov networks.
\newblock \emph{Submitted. Available at http://www. stat. lsa. umich. edu/\~{}
  elevina}, 2010.

\bibitem[H{\"o}fling and Tibshirani(2009)]{hoefling2009}
H.~H{\"o}fling and R.~Tibshirani.
\newblock Estimation of sparse binary pairwise markov networks using
  pseudo-likelihoods.
\newblock \emph{The Journal of Machine Learning Research}, 10:\penalty0
  883--906, 2009.

\bibitem[Jalali et~al.(2011)Jalali, Ravikumar, Vasuki, Sanghavi, ECE, and
  CS]{jalali2011}
A.~Jalali, P.~Ravikumar, V.~Vasuki, S.~Sanghavi, UT~ECE, and UT~CS.
\newblock On learning discrete graphical models using group-sparse
  regularization.
\newblock In \emph{Proceedings of the International Conference on Artificial
  Intelligence and Statistics (AISTATS)}, 2011.

\bibitem[Kim et~al.(2009)Kim, Sohn, and Xing]{kim2009multivariate}
Seyoung Kim, Kyung-Ah Sohn, and Eric~P Xing.
\newblock A multivariate regression approach to association analysis of a
  quantitative trait network.
\newblock \emph{Bioinformatics}, 25\penalty0 (12):\penalty0 i204--i212, 2009.

\bibitem[Koller and Friedman(2009)]{koller2009}
D.~Koller and N.~Friedman.
\newblock \emph{Probabilistic graphical models: principles and techniques}.
\newblock The MIT Press, 2009.

\bibitem[Lauritzen(1996)]{Lauritzen1996}
S.L. Lauritzen.
\newblock \emph{Graphical models}, volume~17.
\newblock Oxford University Press, USA, 1996.

\bibitem[Lee et~al.(2013)Lee, Sun, and Taylor]{lee2013model}
Jason~D Lee, Yuekai Sun, and Jonathan Taylor.
\newblock On model selection consistency of m-estimators with geometrically
  decomposable penalties.
\newblock \emph{arXiv preprint arXiv:1305.7477}, 2013.

\bibitem[Lee et~al.(2012)Lee, Sun, and Saunders]{lee2012proximal}
J.D. Lee, Y.~Sun, and M.A. Saunders.
\newblock Proximal newton-type methods for minimizing convex objective
  functions in composite form.
\newblock \emph{arXiv preprint arXiv:1206.1623}, 2012.

\bibitem[Lee et~al.(2006)Lee, Ganapathi, and Koller]{lee2006}
S.I. Lee, V.~Ganapathi, and D.~Koller.
\newblock Efficient structure learning of markov networks using
  l1regularization.
\newblock In \emph{NIPS}, 2006.

\bibitem[Liang and Jordan(2008)]{liang2008asymptotic}
P.~Liang and M.I. Jordan.
\newblock An asymptotic analysis of generative, discriminative, and
  pseudolikelihood estimators.
\newblock In \emph{Proceedings of the 25th international conference on Machine
  learning}, pages 584--591. ACM, 2008.

\bibitem[Liu and Ihler(2011)]{liu2011learning}
Q.~Liu and A.~Ihler.
\newblock Learning scale free networks by reweighted l1 regularization.
\newblock In \emph{Proceedings of the 14th International Conference on
  Artificial Intelligence and Statistics (AISTATS)}, 2011.

\bibitem[Liu and Ihler(2012)]{liu2012distributed}
Q.~Liu and A.~Ihler.
\newblock Distributed parameter estimation via pseudo-likelihood.
\newblock In \emph{Proceedings of the International Conference on Machine
  Learning (ICML)}, 2012.

\bibitem[Meinshausen and B{\"u}hlmann(2006)]{meinshausen06}
N.~Meinshausen and P.~B{\"u}hlmann.
\newblock High-dimensional graphs and variable selection with the lasso.
\newblock \emph{The Annals of Statistics}, 34\penalty0 (3):\penalty0
  1436--1462, 2006.

\bibitem[Peng et~al.(2009)Peng, Wang, Zhou, and Zhu]{peng2009}
J.~Peng, P.~Wang, N.~Zhou, and J.~Zhu.
\newblock Partial correlation estimation by joint sparse regression models.
\newblock \emph{Journal of the American Statistical Association}, 104\penalty0
  (486):\penalty0 735--746, 2009.

\bibitem[Ravikumar et~al.(2010)Ravikumar, Wainwright, and
  Lafferty]{ravikumar2010}
P.~Ravikumar, M.J. Wainwright, and J.D. Lafferty.
\newblock High-dimensional ising model selection using l1-regularized logistic
  regression.
\newblock \emph{The Annals of Statistics}, 38\penalty0 (3):\penalty0
  1287--1319, 2010.

\bibitem[Rothman et~al.(2010)Rothman, Levina, and Zhu]{rothman2010sparse}
Adam~J Rothman, Elizaveta Levina, and Ji~Zhu.
\newblock Sparse multivariate regression with covariance estimation.
\newblock \emph{Journal of Computational and Graphical Statistics}, 19\penalty0
  (4):\penalty0 947--962, 2010.

\bibitem[Schmidt(2010)]{schmidt2010}
M.~Schmidt.
\newblock \emph{Graphical Model Structure Learning with l1-Regularization}.
\newblock PhD thesis, University of British Columbia, 2010.

\bibitem[Schmidt et~al.(2008)Schmidt, Murphy, Fung, and Rosales]{schmidt2008}
M.~Schmidt, K.~Murphy, G.~Fung, and R.~Rosales.
\newblock Structure learning in random fields for heart motion abnormality
  detection.
\newblock \emph{CVPR. IEEE Computer Society}, 2008.

\bibitem[Schmidt et~al.(2011)Schmidt, Kim, and Sra]{schmidt2011}
M.~Schmidt, D.~Kim, and S.~Sra.
\newblock Projected newton-type methods in machine learning.
\newblock 2011.

\bibitem[Tur and Castelo(2012)]{tur2012learning}
Inma Tur and Robert Castelo.
\newblock Learning mixed graphical models from data with p larger than n.
\newblock \emph{arXiv preprint arXiv:1202.3765}, 2012.

\bibitem[Wainwright and Jordan(2008)]{wainwright2008}
M.J. Wainwright and M.I. Jordan.
\newblock Graphical models, exponential families, and variational inference.
\newblock \emph{Foundations and Trends{\textregistered} in Machine Learning},
  1\penalty0 (1-2):\penalty0 1--305, 2008.

\bibitem[Witten and Tibshirani(2009)]{witten2009covariance}
Daniela~M Witten and Robert Tibshirani.
\newblock Covariance-regularized regression and classification for high
  dimensional problems.
\newblock \emph{Journal of the Royal Statistical Society: Series B (Statistical
  Methodology)}, 71\penalty0 (3):\penalty0 615--636, 2009.

\bibitem[Wright et~al.(2009)Wright, Nowak, and Figueiredo]{wright2009sparse}
S.J. Wright, R.D. Nowak, and M.A.T. Figueiredo.
\newblock Sparse reconstruction by separable approximation.
\newblock \emph{Signal Processing, IEEE Transactions on}, 57\penalty0
  (7):\penalty0 2479--2493, 2009.

\bibitem[Yang et~al.(2012)Yang, Ravikumar, Allen, and Liu]{yang2012graphical}
E.~Yang, P.~Ravikumar, G.~Allen, and Z.~Liu.
\newblock Graphical models via generalized linear models.
\newblock In \emph{Advances in Neural Information Processing Systems 25}, pages
  1367--1375, 2012.

\bibitem[Yang et~al.(2013)Yang, Ravikumar, Allen, and Liu]{yang2013graphical}
E.~Yang, P.~Ravikumar, G.I. Allen, and Z.~Liu.
\newblock On graphical models via univariate exponential family distributions.
\newblock \emph{arXiv preprint arXiv:1301.4183}, 2013.

\bibitem[Yuan and Lin(2006)]{yuan2006model}
M.~Yuan and Y.~Lin.
\newblock Model selection and estimation in regression with grouped variables.
\newblock \emph{Journal of the Royal Statistical Society: Series B (Statistical
  Methodology)}, 68\penalty0 (1):\penalty0 49--67, 2006.

\bibitem[Zhao and Yu(2006)]{zhao2006model}
Peng Zhao and Bin Yu.
\newblock On model selection consistency of lasso.
\newblock \emph{The Journal of Machine Learning Research}, 7:\penalty0
  2541--2563, 2006.

\end{thebibliography}

\section*{Appendix}
\subsection{Proof of Convexity}
\label{app:Proofs}
\noindent {\bf Proposition~\ref{prop:cvx}.}{\em
The negative log pseudolikelihood in  \eqref{eq:negpl} is jointly convex in all the parameters $\{\beta_{ss},\beta_{st}, \alpha_{s}, \phi_{rj}, \rho_{sj}\}$ over the region $\beta_{ss}>0$. 
}

\medskip
\begin{proof}
To verify the convexity of $\tilde{\ell}(\Theta|x,y)$, it suffices to check that each term is convex. 
\noindent$-\log{p(y_r| y_{\backslash r,}, x;\Theta)}$ is jointly convex in $\rho$ and $\phi$ since it is a multiclass logistic regression.
We now check that $-\log{p(x_s | x_{\backslash s}, y;\Theta)}$ is convex. $-\frac{1}{2} \log{\beta_{ss}} $ is a convex function. To establish that $\frac{\beta_{ss}}{2} \left(\frac{\alpha_s}{\beta_{ss}}+\sum_{j}\frac{\rho_{sj}(y_j)}{\beta_{ss}} - \sum_{t\neq s} \frac{\beta_{st}}{\beta_{ss}} x_{t} - x_{s}\right)^2$ is convex, we use the fact that $f(u,v)= \frac{v}{2} (\frac{u}{v} -c)^2$ is convex. Let $v=\beta_{ss}$,  $u= \alpha_s + \sum_{j} \rho_{sj} (y_j) - \sum_{t\neq s} \beta_{st} x_{t}$, and $c= x_s$. Notice that $x_s$, $\alpha_s$, $y_j$, and $x_t$ are fixed quantities and $u$ is affinely related to $\beta_{st}$ and $\rho_{sj}$. A convex function composed with an affine map is still convex, thus $\frac{\beta_{ss}}{2} \left(\frac{\alpha_s}{\beta_{ss}}+\sum_{j}\frac{\rho_{sj}(y_j)}{\beta_{ss}} - \sum_{t\neq s} \frac{\beta_{st}}{\beta_{ss}} x_{t} - x_{s}\right)^2$ is convex.

To finish the proof, we verify that $f(u,v)= \frac{v}{2} (\frac{u}{v} -c)^2 = \frac{1}{2} \frac{(u-cv)^2}{v}$ is convex over $v>0$. The epigraph of a convex function is a convex set iff the function is convex. Thus we establish that the set $C= \{ (u,v,t) |  \frac{1}{2} \frac{(u-cv)^2}{v}\le t, v>0\}$ is convex. Let $
A = \begin{bmatrix}
v&u-cv\\
u-cv&t
\end{bmatrix}.
$
The Schur complement criterion of positive definiteness says $A \succ 0$ iff $v>0$ and $t>\frac{(u-cv)^2}{v}$. The condition $A \succ 0$ is a linear matrix inequality and thus convex in the entries of $A$. The entries of $A$ are linearly related to $u$ and $v$, so $A\succ 0$ is also convex in $u$ and $v$. Therefore $v>0$ and $t>\frac{(u-cv)^2}{v}$ is a convex set.
\end{proof}
\subsection{Sampling From The Joint Distribution}
\label{app:sampling}
In this section we discuss how to draw samples $(x,y) \thicksim p(x,y)$. Using the property that $p(x,y)=p(y) p(x|y)$, we see that if $y \thicksim p(y) $ and $ x \thicksim p(x|y)$ then $(x,y) \thicksim p(x,y)$. We have that 
\begin{align}
p(y) & \propto \exp{(\sum_{r,j} \phi_{rj} (y_r, y_j) +\frac{1}{2}  \rho(y)^{T} B^{-1} \rho(y))}\\
(\rho(y))_s&= \sum_{j} \rho_{sj}(y_j)\\
p(x|y)&= No( B^{-1} (\alpha+\rho(y) ), B^{-1})
\end{align}
The difficult part is to sample $y\thicksim p(y)$ since this involves the partition function of the discrete MRF. This can be done with MCMC for larger models and junction tree algorithm or exact sampling for small models.

\subsection{Maximum Likelihood}
\label{app:mle}
The difficulty in MLE is that in each gradient step we have to compute $\hat{T}(x,y) -E_{p(\Theta)}\left[T(x,y)\right]$, the difference between the empirical sufficient statistic $\hat{T}(x,y)$  and the expected sufficient statistic. In both continuous and discrete graphical models the computationally expensive step is evaluating $E_{p(\Theta)}\left[T(x,y)\right]$. In discrete problems, this involves a sum over the discrete state space and in continuous problem, this requires matrix inversion. For both discrete and continuous models, there has been much work on addressing these difficulties. For discrete models, the junction tree algorithm is an exact method for evaluating marginals and is suitable for models with low tree width. Variational methods such as belief propagation and tree reweighted belief propagation work by optimizing a surrogate likelihood function by approximating the partition function $Z(\Theta)$ by a tractable surrogate $\widetilde{Z}(\Theta)$ \cite{wainwright2008}. In the case of a large discrete state space, these methods can be used to approximate $p(y)$ and do approximate maximum likelihood estimation for the discrete model. Approximate maximum likelihood estimation can also be done via Monte Carlo estimates of the gradients $\hat{T}(x,y) -E_{p(\Theta)}(T(x,y))$.  For continuous Gaussian graphical models, efficient algorithms based on block coordinate descent \cite{glasso,banerjee2008}  have been developed, that do not require matrix inversion.

The joint distribution and loglikelihood are:
\begin{align*}
p(x,y;\Theta)&= \exp{(-\frac{1}{2} x^{T} B x +(\alpha+\rho(y))^{T} x+\sum_{(r,j)}\phi_{rj}(y_r,y_j))}/Z(\Theta)\\
\ell(\Theta)&=\left(\frac{1}{2} x^{T} B x -(\alpha+\rho(y))^{T} x-\sum_{(r,j)}\phi_{rj}(y_r,y_j)\right)\\
&+\log( \sum_{y'} \int{dx \exp{(-\frac{1}{2}x^{T} B x +(\alpha+\rho(y'))^{T} x )}} \exp(\sum_{(r,j)} \phi_{rj}(y'_r,y'_j)) )\\
\end{align*}
The derivative is
\begin{align*}
\frac{\partial \ell}{\partial B} &= \frac{1}{2} x x^{T}+ \frac{\int dx( \sum_{y'} -\frac{1}{2} xx^{T}  \exp(-\frac{1}{2} x^{T} B x +(\alpha+\rho(y))^{T} x +\sum_{(r,j)} \phi_{rj}(y'_r,y'_j)))}{Z(\Theta)}\\
&=\frac{1}{2}xx^{T} + \int \sum_{y'} (-\frac{1}{2} xx^{T} p(x,y';\Theta))\\
&=\frac{1}{2}xx^{T} + \sum_{y'}\int -\frac{1}{2} xx^{T} p(x|y';\Theta) p(y') \\
&=\frac{1}{2}xx^{T} + \sum_{y'}\int -\frac{1}{2} \left(B^{-1} + B^{-1} (\alpha+\rho(y') )(\alpha+ \rho(y')^{T}) B^{-1}\right) p(y') 
\end{align*}
The primary cost is to compute $B^{-1}$ and the sum over the discrete states $y$.
\newline
The computation for the derivatives of $\ell(\Theta)$ with respect to $\rho_{sj}$ and $\phi_{rj}$ are similar.
\begin{align*}
\frac{\partial \ell}{\phi_{rj}(a,b) }&= -1(y_r =a,y_j =b)+\sum_{y'}\int dx 1(y'_r=a,y'_j=b) p(x,y';\Theta)\\
&= -1(y_r =a,y_j =b)+\sum_{y'} 1(y'_r=a,y'_j=b) p(y')
\end{align*}
The gradient requires summing over all discrete states.
\newline
Similarly for $\rho_{sj}(a)$:
\begin{align*}
\frac{\partial \ell}{\rho_{sj}(a) }= -1(y_j = a)x_{s}+\sum_{y'}\int dx (1(y'_j=a) x_s ) p(x',y';\Theta)\\
=-1(y_j = a)x_{s}+\int dx \sum_{y_{\backslash j}'} x_s p(x|y'_{\backslash j},y'_j=a)p(y'_{\backslash j},y'_j = a)
\end{align*}
MLE estimation requires summing over the discrete states to compute the expected sufficient statistics. This may be approximated using using samples $(x,y) \thicksim p(x,y;\Theta)$. The method in the previous section shows that sampling is efficient if $y \thicksim p(y)$ is efficient. This allows us to use  MCMC methods developed for discrete MRF's such as Gibbs sampling.
\subsection{Choosing the Weights}
\label{app:weights}
We first show how to compute $w_{sj}$. 
The gradient of the pseudo-likelihood with respect to a parameter $\rho_{sj} (a) $ is given below
\begin{align}
\frac{\partial \tilde{\ell}}{\partial \rho_{sj}(a)}&= \sum_{i=1}^n  -2\times \indicator{y_{j}^{i}=a} x_{s}^{i}+E_{p_{F}} ( \indicator{y_{j}=a } x_{s}|y_{\backslash j}^{i}, x^{i})  + E_{p_{F}} ( \indicator{y_{j}=a } x_{s}|x^{i}_{\backslash s}, y^{i}) \nonumber \\
&=\sum_{i=1}^n  -2\times \indicator{y_{j}^{i}=a} x_{s}^{i}+x^{i}_{s} p(y_{j}=a)   + \indicator{y_{j}^{i} =a} \mu_{s} \nonumber \\
&= \sum_{i=1}^n \indicator{y_{j}^{i} =a} \left(\hat{\mu}_{s} - x^{i}_{s}\right) +x_{s}^{i} \left( \hat{p}(y_{j} =a) - \indicator {y_{j}^{i}=a} \right) \nonumber\\
&=\sum_{i=1}^n \left(\indicator{y_{j}^{i} =a}-\hat{p}(y_{j}=a) \right) \left(\hat{\mu}_{s} - x^{i}_{s}\right) +\left(x_{s}^{i}-\hat{\mu}_{s}\right) \left( \hat{p}(y_{j} =a) - \indicator {y_{j}^{i}=a} \right)\\
&= \sum_{i=1}^n 2\left(\indicator{y_{j}^{i} =a}-\hat{p}(y_{j}=a) \right) \left(\hat{\mu}_{s} - x^{i}_{s}\right)
\end{align}
Since the subgradient condition includes a variable if $\norm{\frac{\partial \tilde{\ell}}{\partial \rho_{sj}}} > \lambda $, we compute $E\norm{\frac{\partial \tilde{\ell}}{\partial \rho_{sj}}}^{2}$. By independence,
\begin{align}
&E_{p_{F}}\left(\norm{\sum_{i=1}^n 2\left(\indicator{y_{j}^{i} =a}-\hat{p}(y_{j}=a) \right) \left(\hat{\mu}_{s} - x^{i}_{s}\right)}^2 \right )\\
&= 4n E_{p_{F}}\left(\norm{\indicator{y_{j}^{i} =a}-\hat{p}(y_{j}=a)}^2\right) E_{p_{F}}\left( \norm{\hat{\mu}_{s} -x^{i}_{s}}^2\right)\\
&= 4(n-1) p(y_{j}=a) (1-p(y_{j}=a)) \sigma_{s}^{2}
\end{align}
The last line is an equality if we replace the sample means $\hat{p}$ and $\hat{\mu}$ with the true values $p$ and $\mu$. Thus for the entire vector $\rho_{sj}$ we have $E_{p_{F}}\norm{\frac{\partial \tilde{\ell}}{\partial \rho_{sj}}}^{2} =4(n-1) \left(\sum_{a} p(y_{j}=a) (1-p(y_{j}=a)\right) \sigma_{s}^2$. If we let the vector $z$ be the indicator vector of the categorical variable $y_{j}$, and let the vector $p=p(y_{j}=a)$, then $E_{p_{F}}\norm{\frac{\partial \tilde{\ell}}{\partial \rho_{sj}}}^{2} =4(n-1) \sum_{a} p_{a} (1-p_{a}) \sigma^2 = 4(n-1) \mathbf{tr}(\mathbf{cov}(z)) \mathbf{var}(x)$ and $w_{sj} = \sqrt{\sum_{a} p_a (1-p_a ) \sigma_{s}^2}$.

We repeat the computation for $\beta_{st}$. 
\begin{align*}
\frac{\partial \ell}{\partial \beta_{st}} &= \sum_{i=1}^{n}-2 x^{i}_{s} x_{t} +E_{p_F} (x^{i}_{s} x^{i}_{t}|x_{\backslash s},y) +E_{p_F} (x^{i}_{s} x^{i}_{t} |x_{\backslash t},y)\\
&=\sum_{i=1}^{n}-2x^{i}_{s}x^{i}_{t} +\hat{\mu}_{s} x^{i}_{t} +\hat{\mu}_{t} x^{i}_{s}\\
&=\sum_{i=1}^{n} x^{i}_{t}( \hat{\mu_{s}} -x^{i}_{s}) +x^{i}_{s} (\hat{\mu}_{t} -x^{i}_{t})\\
&= \sum_{i=1}^{n}(x^{i}_{t}-\hat{\mu}_{t})(\hat{\mu_{s}} -x^{i}_{s}) +(x^{i}_{s} - \hat{\mu_{s}})(\hat{\mu_{t}} -x^{i}_{t} )\\
&=\sum_{i=1}^{n}2 (x^{i}_{t}-\hat{\mu}_{t})(\hat{\mu_{s}} -x^{i}_{s})
\end{align*}
Thus
\begin{align*}
E&\left( \norm{\sum_{i=1}^{n}2 (x^{i}_{t}-\hat{\mu}_{t})(\hat{\mu_{s}} -x^{i}_{s})}^{2} \right)\\
&=4n E_{p_F} \norm{x_{t} -\hat{\mu_{t}}}^2 E_{p_F} \norm{x_{s} - \hat{\mu}_{s}}^2\\
&= 4(n-1) \sigma_{s}^2 \sigma_{t}^2
\end{align*}
Thus $E_{p_F} \norm{\frac{\partial \ell}{\partial \beta_{st}}}^2 =4(n-1) \sigma_{s}^2 \sigma_{t}^2$ and taking square-roots gives us $w_{st}=\sigma_{s} \sigma_{t}$.\\
We repeat the same computation for $\phi_{rj}$. Let $p_{a} = Pr (y_r =a)$ and $q_{b} =Pr( y_j =b)$. 
\begin{align*}
\frac{\partial \tilde{\ell}}{\partial \phi_{rj}(a,b)} &=
\sum_{i=1}^n -\indicator{y^{i}_{r}=a}\indicator{y^{i}_{j}=b} +E\left(\indicator{y_{r}=a}\indicator{y_{j}=b}|y_{\backslash r},x\right)\\&+E\left(\indicator{y_{r}=a}\indicator{y_{j}=b}|y_{\backslash j}, x\right)\\
&=\sum_{i=1}^n -\indicator{y^{i}_{r}=a}\indicator{y^{i}_{j}=b} +\hat{p}_{a}\indicator{y^{i}_{j}=b}+\hat{q}_{b} \indicator{y^{i}_{r}=a}\\
&=\sum_{i=1}^n\indicator{y^{i}_{j}=b}(\hat{p}_a - \indicator{y^{i}_r =a}) +\indicator{y^{i}_r =a }( \hat{q}_b - \indicator{y^{i}_j =b})\\
&=\sum_{i=1}^n(\indicator{y^{i}_{j}=b}-\hat{q}_b )(\hat{p}_a - \indicator{y^{i}_r =a}) +(\indicator{y^{i}_r =a }-\hat{p}_a )( \hat{q}_b - \indicator{y^{i}_j =b})\\
&= \sum_{i=1}^n 2(\indicator{y^{i}_{j}=b}-\hat{q}_b )(\hat{p}_a - \indicator{y^{i}_r =a})
\end{align*}
Thus we compute 
\begin{align*}
E_{p_F } \norm{\frac{\partial \tilde{\ell}}{\partial \phi_{rj}(a,b)}}^2&=E\left( \norm{\sum_{i=1}^{n} 2(\indicator{y^{i}_{j}=b}-\hat{q}_b )(\hat{p}_a - \indicator{y^{i}_r =a})}^{2} \right)\\
&=4n E_{p_F} \norm{\hat{q}_b - \indicator{y_{j}=b}}^2 E_{p_F} \norm{\hat{p}_a - \indicator{y_{r}=a}}^2\\
&= 4(n-1) q_b (1-q_b ) p_a (1-p_a )
\end{align*}
From this, we see that $E_{p_F } \norm{\frac{\partial \tilde{\ell}}{\partial \phi_{rj}}}^2 = \sum_{a=1}^{L_r} \sum_{b=1}^{L_j} 4(n-1) q_b (1- q_b ) p_{a} (1- p_a )$ and $w_{rj}=\sqrt{\sum_{a=1}^{L_r} \sum_{b=1}^{L_j}  q_b (1- q_b ) p_{a} (1- p_a )}$.


\end{document}